\documentclass[letterpaper]{article}
\usepackage{aaai}
\usepackage{iftex}
\ifPDFTeX
  \usepackage{times}
  \usepackage{helvet}
  \usepackage{courier}
\else
  \usepackage{fontspec}
\fi
\usepackage{amsmath}
\usepackage{amssymb}
\usepackage{amsfonts}
\usepackage{booktabs}
\usepackage{multirow}
\usepackage{makecell}
\usepackage{graphicx}
\usepackage{xcolor}
\usepackage{colortbl}
\usepackage{tikz}
\usepackage{pifont}
\usepackage{enumitem}
\usepackage{tabularx}
\usepackage{graphicx}
\usepackage{subcaption}
\usepackage{placeins}
\usepackage{float}
\usepackage{listings}
\usepackage{url}
\usetikzlibrary{arrows.meta,positioning,fit,calc,shapes.geometric}

\ifdefined
\fi

\definecolor{seablue}{RGB}{43,92,130}
\definecolor{softblue}{RGB}{232,241,248}
\definecolor{softgray}{RGB}{244,246,248}
\definecolor{softgreen}{RGB}{231,244,237}
\definecolor{softorange}{RGB}{253,239,219}
\definecolor{linegray}{RGB}{110,122,132}
\definecolor{bestblue}{RGB}{236,244,252}

\lstdefinestyle{appendixlisting}{
  basicstyle=\ttfamily\tiny,
  breaklines=true,
  breakatwhitespace=false,
  columns=fullflexible,
  keepspaces=true,
  showstringspaces=false,
  frame=single,
  framerule=0.2pt,
  xleftmargin=0pt,
  xrightmargin=0pt
}

\newcommand{\method}{\textsc{AlphaSchema}}
\newcommand{\cmark}{\ding{51}}
\newcommand{\xmark}{\ding{55}}
\newcommand{\best}[1]{\textbf{#1}}
\newcommand{\second}[1]{\underline{#1}}

\nocopyright
\title{AlphaSchema: Exploring the Space of Trading\\
Semantics for LLM-Based Alpha Mining%
\thanks{Preprint. Correspondence: \texttt{yijingyang@x-tech.net.cn} and \texttt{jyan0311@student.monash.edu}}%
\thanks{\protect\url{https://github.com/JingyangYi/AlphaSchema}}}
\author{
{\Large\bfseries Jingyang Yi$^{1,2}$ \quad
Jian Yang$^{2,3}$ \quad
Yifei Jin$^{1,2}$}\\[0.35em]
{\Large\bfseries Yuqi Li$^{2,4}$ \quad
Jian Li$^{5}$}\\[0.75em]
{\normalfont\normalsize $^{1}$X-Tech \quad
$^{2}$Xtech-PandaAI-Waton Joint Lab}\\
{\normalfont\normalsize $^{3}$Monash University \quad
$^{4}$PandaAI \quad
$^{5}$IIIS, Tsinghua University}
}

\begin{document}
\maketitle

\begin{abstract}
Automated alpha mining has increasingly adopted large language model
(LLM) agents for factor generation and iterative discovery. However,
existing LLM-based systems often delegate both factor construction and
search decisions to the agent itself, without an explicit exploration
space or a principled mechanism for navigating that space. As a result, exploration remains largely implicit and difficult to control or optimize
systematically. We introduce \method{}, which constructs and explores a structured space
of trading semantics for alpha mining. Each point in this space is a
schema plan composed of \emph{Event}, \emph{Context}, \emph{Qualities},
\emph{Direction}, and \emph{Output}, specifying the semantics of a
candidate factor before implementation. \method{} decouples exploration
from implementation: an LLM translates selected schema plans into
executable factors, while evaluated rewards are accumulated to learn a
surrogate model over the semantic space. An iterative selection mechanism uses
this model to balance global exploration, surrogate-guided exploitation,
and local mutation. Experiments on the Chinese stock market show that \method{} discovers
factor pools with strong predictive and portfolio performance. Further analyses show that the semantic search process navigates diverse regions while increasingly allocating evaluations toward high-reward regions, and that implementations of the same schema plans by different LLMs
exhibit comparable predictive quality, suggesting that alpha mining quality is largely robust to the choice of LLM within our framework.
\end{abstract}

\begin{figure*}[t]
    \centering
    \includegraphics[width=0.98\textwidth]{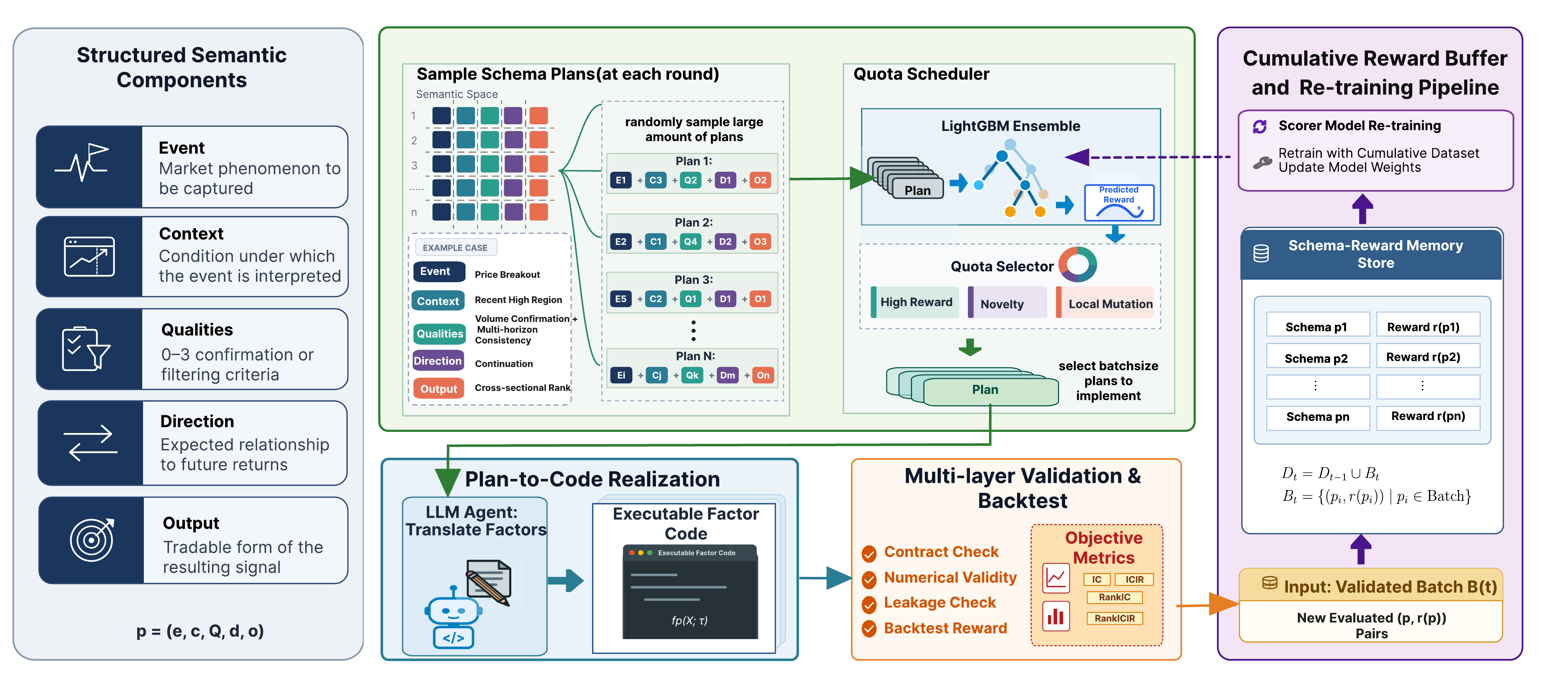}
    \caption{Overview of \method{}. \method{} performs iterative alpha mining
    over a structured trading-semantic space. At each round, semantic plans are
    generated and selected by a quota-based semantic scheduler that allocates
    the evaluation budget among structural exploration, surrogate-guided
    exploitation, and local mutation. Selected plans are realized into
    executable factors by an LLM agent and evaluated through guarded
    backtesting. The resulting plan--reward pairs are accumulated to update the
    semantic scorer, which guides subsequent search rounds.}
    \label{fig:method_overview}
\end{figure*}

\section{Introduction}
\label{sec:introduction}

Alpha mining aims to discover trading signals that are predictive, executable, interpretable, and stable out of sample. Traditional asset-pricing research derives factors from economic hypotheses and
empirical anomalies \cite{sharpe1964capital,fama1992cross,fama1993common}, but the vast and evolving space of
potential trading mechanisms makes manual discovery inherently limited. An automated factor-mining system should therefore not only generate candidate signals, but also efficiently explore meaningful trading hypotheses while preserving economic interpretation, implementation validity, and reproducible evaluation.

Existing automated systems have expanded alpha discovery through formulaic and evolutionary search. Formulaic and evolutionary methods represent factors as expressions, syntax trees, or operator programs \cite{kakushadze2016101,koza1994genetic,zhaofan2022genetic,zhang2020autoalpha,cui2021alphaevolve,yu2023generating,shi2025alphaforge}, enabling large-scale exploration of implementation spaces. However, the search process is primarily driven by mathematical composition rather than explicit economic semantics, making it difficult to
incorporate higher-level trading hypotheses into the discovery process. 

More recent LLM-based agents introduce planning, code generation, memory systems, trajectory evolution, and guarded execution into factor discovery \cite{li2024can,tang2025alphaagent,shi2026navigating,lin2026factorengine,han2026quantaalpha,shi2026hubble,liu2026cognitive,chen2025alphasage}. However, many existing LLM-based systems couple search-space construction, search-trajectory selection, and code-level realization inside the LLM agent. Consequently, the candidate space is shaped implicitly by prompts, memory traces, and model-specific proposal behavior, making diversity difficult to measure, coverage difficult to control, and search trajectories difficult to reproduce or improve systematically. Moreover, the performance of such paradigms remains tightly coupled with the capabilities of frontier LLMs, incurring substantial computational costs and limiting the scalability of large-scale automated mining.

We propose \method{}, a semantic-space exploration framework for automated alpha mining that separates \emph{what to search} from \emph{how to implement it}. Instead of treating code or formulas as the primary search object, \method{} represents each candidate as a structured trading-semantic plan
\begin{equation}
p=(e,c,Q,d,o),
\end{equation}
where \(e\) denotes a market event, \(c\) its conditioning context, \(Q\) a set of quality constraints, \(d\) the directional hypothesis, and \(o\) the output form. The search process operates over this schema space, while an LLM translates selected plans into executable factor implementations for evaluation.

\method{} maintains a reward buffer of evaluated plans and trains a LightGBM reward model over schema features \cite{ke2017lightgbm}. At each round, a large sampled candidate pool is scored before LLM implementation. A quota selector allocates the limited implementation and backtesting budget across structurally novel candidates, surrogate-ranked high-reward candidates, and mutation-based refinements around historically strong plans. By associating evaluation feedback with schema plans rather than generated code or operators, \method{} learns reusable priors over which trading semantics are worth exploring.

The main contributions of this work are summarized as follows:
\begin{itemize}[leftmargin=12pt]
    \item We introduce structured trading-semantic plans as the new search
    abstraction for automated alpha mining. Unlike formula- or program-level
    mining, \method{} explores an explicit semantic space of market events,
    contexts, quality constraints, directions, and output forms before code
    realization.
    \item We develop a reward-guided semantic exploration algorithm for
    LLM-based alpha mining. Using accumulated plan-level rewards and adaptive
    selection, \method{} makes the search space explicit and optimizable,
    reducing reliance on implicit exploration behaviors of individual LLM
    agents.
    \item We empirically validate the proposed semantic search framework.
    Experiments show strong predictive and portfolio performance of discovered
    factors, while further analyses demonstrate that semantic plans provide
    learnable search signals, support effective exploration, and remain robust
    across different LLM realizations.
\end{itemize}

\section{Preliminary}
\label{sec:preliminary}

\subsection{Alpha Mining}

An alpha factor is a rule that transforms historical market information
into a cross-sectional signal for asset ranking or prediction. Given
\(N\) tradable assets observed over \(T\) trading dates, let
\(X_t\in\mathbb{R}^{N\times D}\) denote the market state at date \(t\),
where rows correspond to assets and columns correspond to observable
market features. A factor \(f\) maps a historical window of market
states into a signal vector:

\[
v_t=f(X_{t-\tau+1:t}),\quad v_t\in\mathbb{R}^{N}.
\]

Beyond its executable implementation, a factor typically reflects an
underlying market mechanism that provides an interpretable rationale
for its predictive relationship with future returns. Therefore, alpha
mining involves not only finding predictive transformations, but also
discovering meaningful mechanisms and realizing them as executable
factors.

Formally, we distinguish a semantic factor plan from its executable
realization. Let \(p\in\mathcal{P}\) denote a candidate trading-semantic
plan, and let

\[
\Gamma:\mathcal{P}\rightarrow\Delta(\mathcal{F})
\]

denote a realization process that maps a semantic plan to a distribution
over executable factors. A realized factor \(f\sim \Gamma(p)\) is evaluated by
a factor-quality objective \(\mathcal{J}(f)\). The objective of semantic
factor discovery can therefore be written as

\[
p^*=\arg\max_{p\in\mathcal{P}}
\mathbb{E}_{f\sim \Gamma(p)}[\mathcal{J}(f)].
\]

A plan does not uniquely determine an executable factor: different
realizations may adopt different operators, parameters, and functional
forms. Thus, the objective concerns the reward distribution induced by a plan rather than a deterministic mapping, and each observed reward provides only a sample of this distribution. In practice, however, evaluating this distribution requires a trade-off between reward estimation accuracy and search coverage under a limited mining budget.

\subsection{Structured Semantic Plans}

To search over trading semantics, \method{} represents each candidate
factor as a structured semantic plan.\footnote{See Appendix~\ref{sec:appendix_schema_construction} for the
schema-vocabulary construction, JSON representation, and a concrete
plan-to-code example.} Rather than assuming a unique
ontology of market mechanisms, we define a practical and expressive
representation that captures recurring dimensions of factor
construction. A semantic plan describes the intended trading mechanism
before implementation and can subsequently be translated into an
executable factor by an LLM.

Formally, a semantic plan is represented as

\[
p=(e,c,Q,d,o)\in\mathcal{P},
\]

where the five dimensions jointly characterize different aspects of a
candidate trading mechanism:

\begin{itemize}[leftmargin=12pt, nosep]
  \item \textbf{Event}: the market phenomenon or signal-generating event
        being captured, such as breakout, volume expansion, or volatility
        compression;

  \item \textbf{Context}: the market state or reference condition under
        which the event is interpreted, such as a recent extreme, VWAP
        region, volatility regime, or cross-sectional quantile;

  \item \textbf{Qualities}: additional properties or validation criteria
        that refine the mechanism, such as volume confirmation,
        multi-horizon consistency, or outlier filtering;

  \item \textbf{Direction}: the expected relationship between the
        mechanism and future returns, such as continuation, reversal, or
        range oscillation;

  \item \textbf{Output}: the form in which the mechanism is expressed as
        a tradable signal, such as a continuous score, event decay, or
        cross-sectional rank.
\end{itemize}

Together, these dimensions specify what market phenomenon is considered,
under which conditions it is interpreted, what evidence supports it, how
it is expected to affect future returns, and how the resulting signal is
expressed. Unlike formula trees or sequential decision processes, a
semantic plan does not impose a predefined ordering or compositional
structure among these dimensions. They jointly define a point in the
trading-semantic space, allowing the search process to explore different
combinations of trading concepts without assuming a fixed construction
procedure.

\section{Methodology}
\label{sec:methodology}

Given the trading-semantic space defined in
Section~\ref{sec:preliminary}, \method{} performs iterative search under
a fixed generation and backtesting budget. In each round, candidate
plans are proposed and selected using predicted reward, structural
novelty, and local mutation. Selected plans are implemented as executable
factors, validated, and backtested. Their corresponding rewards are added to a buffer
to update the semantic surrogate for the next round. 

The search loop has three components:
(1) \emph{semantic plan proposal and selection}, which combines global
sampling, local mutation, and quota-based allocation;
(2) \emph{plan realization and evaluation}, which generates executable
factors and applies execution, leakage, and backtesting checks; and
(3) \emph{semantic reward learning}, which learns from accumulated
plan--reward observations to score sampled candidate plans. Figure~\ref{fig:method_overview} summarizes the closed-loop workflow.

\subsection{Semantic Plan Space Construction}

We instantiate a trading-semantic space using configurable vocabularies
for events, contexts, qualities, directions, and outputs. Let these
vocabularies be denoted by
\(\mathcal{V}_{E}\), \(\mathcal{V}_{C}\), \(\mathcal{V}_{Q}\),
\(\mathcal{V}_{D}\), and \(\mathcal{V}_{O}\), respectively. The resulting
plan space is

\[
\mathcal{P}
=
\mathcal{V}_{E}
\times
\mathcal{V}_{C}
\times
\mathcal{Q}_{\mathrm{set}}
\times
\mathcal{V}_{D}
\times
\mathcal{V}_{O}.
\]

Here
\[
\mathcal{Q}_{\mathrm{set}}
=
\{Q\subseteq\mathcal{V}_{Q}:0\leq |Q|\leq 3\}
\]
denotes the admissible sets of quality constraints. The quality position may
therefore contain zero to three complementary constraints, whereas the
remaining four positions select one primary schema item each.

The vocabularies can be derived from domain expertise, academic and
practitioner literature, existing factor libraries, and findings from
previous discovery runs. They need not cover all trading semantics and
can instead be configured for a target domain, such as equity
fundamentals or futures term structure.

A plan describes only trading logic, without prescribing operators,
library functions, or numerical parameters. We also impose no
hand-crafted compatibility rules across dimensions. Unusual or
potentially inconsistent combinations are evaluated through code
realization and empirical feedback, avoiding complex prior constraints
while retaining unconventional combinations that may still produce
useful factors.

\subsection{Plan Realization and Execution Validation}

Given a selected plan \(p\), the code agent receives its semantic
specification, data contract, and implementation constraints, and
translates it into executable factor functions. The plan specifies only
the intended trading logic; the agent independently chooses the
operators, functional form, and numerical parameters. It implements each
plan at a fast and a slow time scale, yielding realizations
\(f_{p,\mathrm{fast}}\) and \(f_{p,\mathrm{slow}}\).

Before backtesting, each realization passes through execution guards
that verify contract compliance and computational validity, test
numerical stability, and screen for potential look-ahead leakage.
Invalid implementations receive one repair attempt; if they still fail, the
plan is assigned zero reward, using unrecoverable realization failure as
negative semantic feedback.\footnote{The zero reward is a
realization-feasibility penalty rather than a claim that the plan has zero
latent alpha. Keeping failed plans in the buffer helps the surrogate avoid
repeatedly selecting semantic structures that are difficult to implement as
valid, robust factors.}

\subsection{Semantic Reward Modeling}

For each valid realization \(f_{p,s}\) of plan \(p\), where
\(s\in\{\mathrm{fast},\mathrm{slow}\}\), we define
\begin{equation}
\begin{aligned}
r_{p,s}
&=
\alpha\,\mathrm{RankIC}(f_{p,s})
+\beta\,\mathrm{RankICIR}(f_{p,s}) \\
&\quad
-\lambda\,\Delta_{\mathrm{lag}}(f_{p,s}),
\end{aligned}
\label{eq:realization_reward}
\end{equation}
where
\[
\Delta_{\mathrm{lag}}(f)
=
\max\!\left\{
0,\,
\mathrm{RankIC}(f)-\mathrm{RankIC}(L_1f)
\right\},
\]
and \(L_1f\) denotes the signal delayed by one period. The coefficients
\(\alpha,\beta,\lambda>0\) control predictive strength, temporal
stability, and lag sensitivity, respectively; their experimental values are
reported in Appendix~\ref{sec:appendix_search_config}. The plan reward is
\[
r(p)=\max_{s\in\{\mathrm{fast},\mathrm{slow}\}} r_{p,s}.
\]

After each round, new plan--reward pairs are added to the reward buffer\footnote{
We use an accumulated buffer because each realized plan provides expensive
schema-level supervision; retaining all observations improves surrogate
accuracy and later exploitation. Appendix~\ref{sec:appendix_accumulated_buffer}
discusses this choice.}
\[
\mathcal{D}_t
=
\mathcal{D}_{t-1}
\cup
\{(p_i,r(p_i))\}_{i\in\mathcal{B}_t}.
\]
A LightGBM reward model is retrained on the expanded buffer using structured
plan features \(\phi(p)\), which encodes the selected schemas with one-hot features, category
and scope counts, and pairwise schema-component interactions. In each subsequent round, we sample a large
candidate pool from the schema space and assign each sampled plan a predicted
score \(\hat r_t(p)\).

\subsection{Adaptive Quota-Based Plan Selection}

Let \(B\) denote the number of plans evaluated in each search round. We
allocate this batch across exploration, surrogate-guided exploitation,
and local mutation. If \(n_t\) plans have been evaluated before round
\(t\), the exploration share is

\[
\rho_t
=
\rho_{\min}
+
(\rho_{\max}-\rho_{\min})
\exp\!\left(-\frac{n_t}{\tau}\right),
\]

where \(0 \leq \rho_{\min} \leq \rho_{\max} \leq 1\) and \(\tau>0\).
As more plans are evaluated, \(\rho_t\) gradually decreases from
\(\rho_{\max}\) toward \(\rho_{\min}\). In practice, we precede this
schedule with a short cold-start period in which the full batch is
allocated to exploration. After the cold start, a fraction \(\rho_t\) of each batch
is allocated to exploration. The remaining fraction \(1-\rho_t\) is
divided between exploitation and mutation.

Both exploration and surrogate-guided exploitation select plans from the
sampled candidate pool: the former favors structural novelty, while the latter
ranks candidates by \(\hat r_t(p)\).\footnote{Appendix~\ref{sec:appendix_explore_mutation} gives the exact coverage-novelty score and local mutation operator used by the selector.}
Mutation starts from evaluated high-reward plans and changes one schema
component to generate unseen neighbors, providing a simple way to search
locally around promising regions. This local search is particularly relevant in alpha mining, where the
objective is to identify a small set of strong factors.

\section{Experiments}
\label{sec:experiment}

This section first describes the experimental setup, including the dataset,
evaluation protocol, baselines, and search configuration. We then report the
downstream backtesting performance of the factor pools discovered by
\method{}, comparing them with representative predictive models, open-source
factor libraries, and agentic mining systems.

\subsection{Experimental Setup}

\paragraph{Dataset and Metrics.}
Following recent Qlib-based alpha-mining studies, we use the CSI300
universe as the primary benchmark. The data are split into training
(Jan 1, 2016 -- Dec 31, 2020), validation (Jan 1, 2021 -- Dec 31, 2022),
and testing (Jan 1, 2023 -- Dec 31, 2025). The prediction target is the
5-day forward close-to-close return. We report IC, ICIR, RankIC, and
RankICIR for predictive performance, together with benchmark-relative
AER, IR, and MDD for portfolio evaluation. Data processing, trading
costs, and metric definitions are provided in
Appendix~\ref{sec:appendix_eval_details}.

\paragraph{Baselines.}
We compare against four baseline families. Machine-learning predictors
include MLP and XGBoost \cite{chen2016xgboost}. Deep sequence models
include Transformer \cite{vaswani2017attention}, GRU
\cite{cho2014learning}, and LSTM \cite{graves2012long}.
Open-source factor libraries include Alpha158 and Alpha360
\cite{yang2020qlib}. Agentic mining systems include RD-Agent
\cite{li2026r} and QuantaAlpha \cite{han2026quantaalpha}.
Predictor baselines use the same
train/validation/test split.

\paragraph{Implementation Details.}
\method{} searches over a price-volume semantic space consisting of 140
components, including 40 Events, 40 Contexts, 50 Qualities, 3 Directions,
and 7 Outputs.\footnote{Appendix~\ref{sec:appendix_schema_examples} provides examples of schema records from each category.} These components describe trading semantics derived from
OHLCV and VWAP information, while additional fundamental schemas are
introduced separately in the \textsc{+Fundamental} setting. DeepSeek-V4-Flash
is used as the default code-agent backend due to its low inference cost.
Each run evaluates 16 plans per round for 80 rounds, with the first 10
rounds using exploration only and later rounds gradually introducing
surrogate-guided selection and mutation. We conduct five independent runs
and construct final factor pools from validated outputs using reward
ranking and correlation filtering.

\subsection{Main Results}

\subsubsection{Pool-Level Evaluation}

\begin{table*}[t]
\centering
\caption{Main-result table on CSI300. ``Fund. Schema'' indicates whether the mining process is allowed to use fundamental semantic schema fields. Higher is better for all metrics except MDD. Best results are bolded and second-best results are underlined.}
\label{tab:main_eval}
\setlength\tabcolsep{3.0pt}
\small
\resizebox{0.98\textwidth}{!}{
\begin{tabular}{llcccccccc}
\toprule
\multirow{2}{*}{\textbf{Category}} & \multirow{2}{*}{\textbf{Method}} &
\multirow{2}{*}{\makecell{\textbf{Fund.}\\\textbf{Schema}}} &
\multicolumn{4}{c}{\textbf{Factor Predictive Power}} &
\multicolumn{3}{c}{\textbf{Strategy Performance}} \\
\cmidrule(lr){4-7}\cmidrule(lr){8-10}
 & & &
\textbf{IC} & \textbf{ICIR} & \textbf{Rank IC} & \textbf{Rank ICIR} &
\textbf{IR} & \textbf{AER (\%)} & \textbf{MDD (\%)$\downarrow$} \\
\midrule
\rowcolor[RGB]{242,244,247}\multicolumn{10}{c}{\textit{\textbf{Machine-Learning Predictors}}} \\
\multirow{2}{*}{\makecell[l]{Machine\\Learning}}
& MLP & \xmark & 0.0236 & 0.1408 & 0.0389 & 0.2471 & 0.5901 & 5.16 & 11.47 \\
& XGBoost & \xmark & 0.0284 & 0.2024 & 0.0373 & 0.2699 & 0.3556 & 2.55 & \second{10.30} \\
\midrule
\rowcolor[RGB]{242,244,247}\multicolumn{10}{c}{\textit{\textbf{Deep Sequence Models}}} \\
\multirow{3}{*}{\makecell[l]{Deep\\Learning}}
& Transformer & \xmark & 0.0289 & 0.1561 & 0.0507 & 0.2775 & 0.6929 & 6.14 & 12.03\\
& GRU & \xmark & 0.0310 & 0.1782 & \second{0.0565} & \second{0.3242} & 0.4609 & 3.73 & 13.21 \\
& LSTM & \xmark & \second{0.0380} & 0.2269 & \best{0.0587} & \best{0.3521} & 0.6317 & 5.14 & 11.51 \\
\midrule
\rowcolor[RGB]{242,244,247}\multicolumn{10}{c}{\textit{\textbf{Open-Source Factor Libraries}}} \\
\multirow{2}{*}{\makecell[l]{Factor\\Library}}
& Alpha158 & \xmark & 0.0347 & 0.2081 & 0.0504 & 0.3009 & 0.5474 & 4.36 & 15.87 \\
& Alpha360 & \xmark & 0.0231 & 0.1710 & 0.0292 & 0.2100 & 0.4024 & 2.99 & \best{8.86} \\
\midrule
\rowcolor[RGB]{242,244,247}\multicolumn{10}{c}{\textit{\textbf{Agentic Mining Systems}}} \\
\multirow{2}{*}{\makecell[l]{Agentic\\Mining}}
& RD-Agent & \xmark & 0.0242 & 0.1494 & 0.0504 &  0.3094 & \second{0.9861} & 6.81 & 15.57 \\
& QuantaAlpha & \xmark & 0.0208 & 0.1619 & 0.0380 & 0.2934 & 0.6726 & 5.57 & 14.18 \\
\midrule
Ours
& \method{} (OHLCV) & \xmark & \best{0.0382} & \best{0.2374} & 0.0498 & 0.2912 & 0.7624 & \second{8.53} & 18.63 \\
\rowcolor{bestblue}
& \textbf{\method{} (+Fund.)} & \cmark & \second{0.0380} & \second{0.2365} & 0.0487 & 0.2857 & \best{1.0877} & \best{11.94} & 15.43 \\
\bottomrule
\end{tabular}
}
\end{table*}

The principal evaluation of \method{} is conducted at the factor-pool
level. Table~\ref{tab:main_eval} reports predictive and portfolio
performance on the held-out test period.\footnote{Appendix~\ref{sec:appendix_csi500}
reports an additional CSI500 backtest using an independently mined and selected
factor pool under the same downstream protocol.} The OHLCV setting contains 120
discovered factors and achieves the strongest IC (0.0382) and ICIR
(0.2374) among the compared methods.

We also report a \textsc{+Fundamental} variant, which augments the OHLCV
pool with 30 factors whose schema plans include fundamental fields. This
expanded 150-factor pool further improves portfolio performance,
achieving the strongest IR (1.0877) and AER (11.94\%). LSTM records the
highest Rank IC and Rank ICIR, while \method{} performs best on several
other predictive and portfolio metrics. Overall, the results show that
semantic-plan search can discover useful factors and produce competitive
downstream performance when the resulting factors are combined.

\begin{figure}[!t]
    \centering
    \includegraphics[width=\columnwidth]{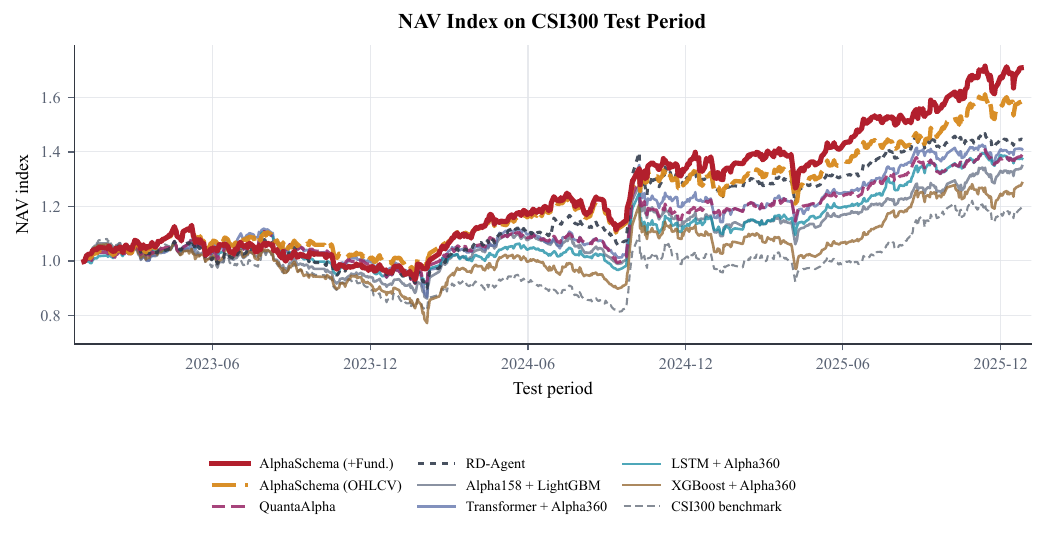}
    \caption{Held-out NAV trajectories against representative baselines and
    the CSI300 benchmark over the 2023--2025 test period. All learned factor
    pools use the same CSRankNorm, LightGBM combiner, and Top50/Drop5 backtest
    protocol.}
    \label{fig:nav}
\end{figure}

Figure~\ref{fig:nav} provides a complementary library-level view of the
same held-out period. It compares the net asset value (NAV) trajectories of
the exported \method{} factor pools with representative predictive models,
agentic mining systems, open-source factor libraries, and the CSI300
benchmark under the same backtest setting. The \method{} curves maintain
higher NAV growth through the test window, indicating that the selected
factor pools convert predictive signal into stronger realized portfolio
value.

\section{Analysis}
\label{sec:analysis}

Beyond overall performance, we further analyze why \method{} is
effective as a semantic search framework for alpha mining. Our analyses
focus on four aspects: (1) how the proposed schema decomposition contributes
semantic structure to factor discovery; (2) how search redistributes
evaluations across the resulting semantic space; (3) how realization budget
should be allocated under stochastic plan-to-factor mapping; and (4) how
implementation robustness changes across different LLMs.
\subsection{Semantic Component Ablation}

We first examine how individual schema components contribute useful
information to factor discovery. Unless otherwise noted, all variants
share the same LLM backend and evaluation protocol. 

We select 100 complete schema plans and construct five
leave-one-component-out variants by removing one field from
\(\{\emph{Event}, \emph{Context}, \emph{Qualities}, \emph{Direction},
\emph{Output}\}\). Each incomplete plan is provided to DeepSeek-V4-Flash
for factor realization, with one repair opportunity allowed under the
same execution protocol. The generated factors are then evaluated after
execution validation and backtesting.

\begin{figure}[!t]
    \centering
    \includegraphics[width=\columnwidth]{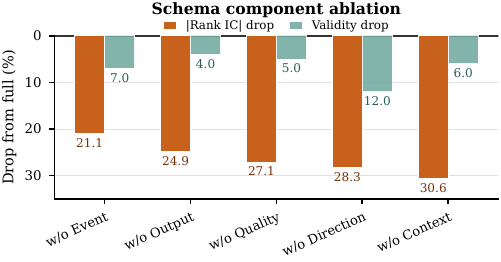}
    \caption{Component-wise schema ablation.}
    \label{fig:schema_leave_one_out}
\end{figure}

Figure~\ref{fig:schema_leave_one_out} summarizes the results. Removing
any schema component leads to a degradation in factor quality, while
the implementation success rate decreases only moderately. Here,
successful implementation requires passing execution checks including
data-contract compliance, numerical validity, and look-ahead leakage
screening. The relatively small drop in validity suggests that the LLM
can still construct executable factors even from incomplete semantic
descriptions\footnote{See Appendix~\ref{sec:appendix_ablation} for further discussion of ablated-schema executability.}, but the missing components reduce the quality of the
resulting factors.

Across all leave-one-out variants, removing a single semantic component
causes a clear reduction in Rank IC compared with the complete schema
plans. This indicates that the five schema dimensions provide
complementary information for factor discovery: although the realization
model can compensate for missing semantic details, the resulting factors
are less effective when important aspects of the trading mechanism are
omitted.

\subsection{Semantic Navigation During Search}
\begin{figure}[!t]
    \centering
    \includegraphics[width=\columnwidth]{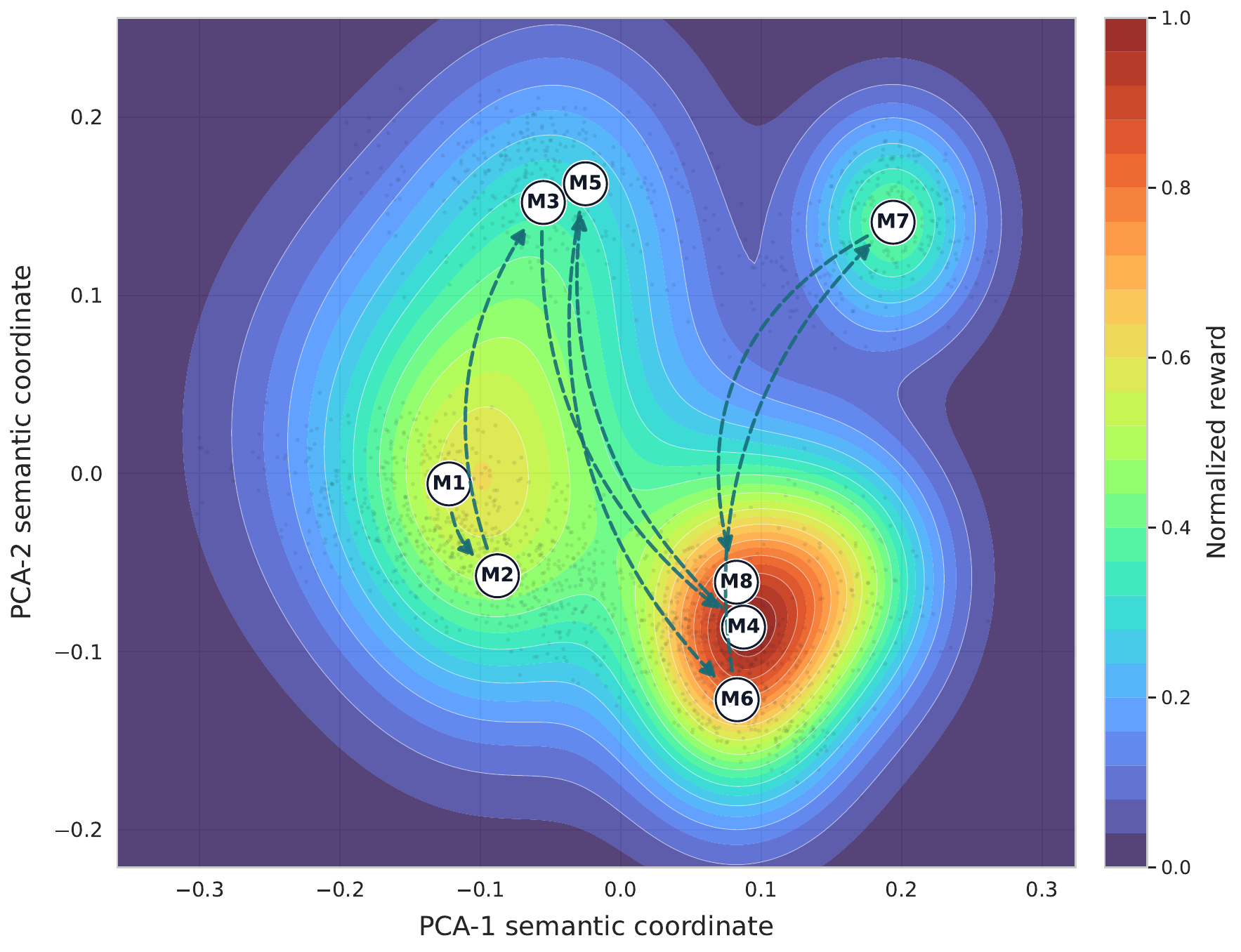}
    \caption{Semantic-space navigation across search rounds.}
    \label{fig:semantic_trajectory}
\end{figure}

To examine how \method{} navigates the semantic space, we trace all
schema plans generated during a complete search run. Each plan is
embedded from its semantic description and reduced to a 50-dimensional
representation using PCA. We first cluster all plans in this space using
K-means to identify global semantic regions. For each search stage, we
identify the dominant region based on the distribution of generated
plans, with temporal smoothing applied to obtain a stable region
trajectory. Figure~\ref{fig:semantic_trajectory} visualizes all plans in
the first two principal components, with arrows connecting the dominant
regions across search stages.\footnote{See Appendix~\ref{sec:appendix_semantic_navigation} for embedding and clustering details, round-level dominance definitions, and the schema interpretation of dominant manifolds.}

The search process initially covers diverse semantic regions and later
allocates more evaluations toward high-reward regions, while maintaining
multiple active areas of exploration. This pattern indicates that the
selector reallocates budget toward reward-enriched semantic regions without
collapsing the search to a single narrow cluster.

\subsection{Realization Variance and Budget Efficiency}

A semantic plan induces a distribution of executable realizations rather
than a deterministic factor. We study the trade-off between estimating
plan quality more accurately and discovering high-quality plans under a
limited realization budget. For each plan, repeated realizations provide
a better estimate of its expected quality, but consume budget that could
otherwise explore additional semantic candidates.

\begin{figure}[!t]
    \centering
    \includegraphics[width=\columnwidth]{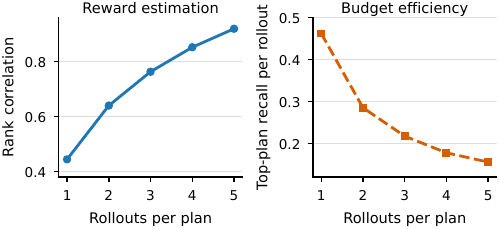}
    \caption{Plan quality estimation and top-plan discovery efficiency.}
    \label{fig:realization_efficiency}
\end{figure}

Using 100 schema plans with up to eight realizations, we treat the average
reward across realizations as an approximate reference quality. Figure
\ref{fig:realization_efficiency} shows that increasing the number of
realizations improves ranking consistency with this reference. However,
when normalized by the number of realizations, the efficiency of
identifying top-quality plans decreases substantially. These results show that repeated realizations improve plan-quality
estimation, but with diminishing returns for identifying top-quality
plans. Thus, single realization provides a more efficient estimator for
large-scale alpha mining, where the objective is ranking promising
candidates rather than accurately estimating every plan's expected
reward.\footnote{See Appendix~\ref{sec:appendix_realization_budget_efficiency}
for the repeated-realization bootstrap and
Appendix~\ref{sec:appendix_single_realization_predictability} for the
schema-level predictability argument. A realized reward is a noisy sample of
plan quality: repeated rollouts reduce noise for one plan, whereas evaluating
more plans exposes more schema combinations. Since schema features recur across
plans, the surrogate can pool noisy single-realization rewards at the feature
level.}

\subsection{LLM Realization Robustness}

Many LLM-based alpha-mining pipelines rely on strong frontier models
because the same model often performs proposal, reflection, selection,
and factor implementation \cite{tang2025alphaagent,shi2026navigating,lin2026factorengine,han2026quantaalpha,shi2026hubble,liu2026cognitive,chen2025alphasage}.
We therefore examine whether \method{}'s discovered factor quality depends
on the LLM used for realization.

\begin{figure}[!t]
    \centering
    \includegraphics[width=\columnwidth]{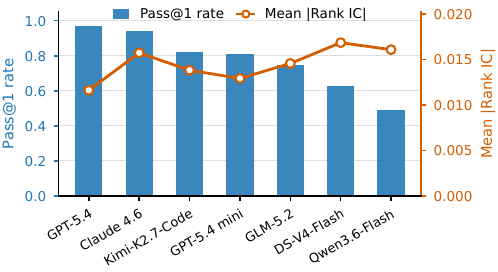}
    \caption{Effect of LLM realization models on implementation success and factor quality.}
    \label{fig:code_agent_model_invariance}
\end{figure}

Holding 100 schema plans fixed, seven LLM backends generate factor
implementations under the same prompt and validation protocol. As shown
in Figure~\ref{fig:code_agent_model_invariance}, Pass@1 varies
substantially across models, indicating different levels of instruction
following and implementation reliability.\footnote{Pass@1 is measured
before repair and therefore reflects each model's one-shot task-completion
ability. The reported Rank IC statistics are computed after allowing one
additional repair attempt when available, so they measure the quality of
successfully realized factors under the full realization protocol.}
However, among successful realizations, mean \(|\mathrm{RankIC}|\) remains
in a narrow range (0.0116--0.0168), with no monotonic relationship to model
capability. This suggests that stronger LLMs mainly improve realization
success, while factor quality is primarily determined by the semantic plan
being implemented.

\section{Related Work}
\label{sec:related_work}

Automated alpha mining has evolved from human-crafted factors and
symbolic heuristics toward large-scale search. Early studies formalized
interpretable return predictors through factor models and anomaly
research \cite{fama1992cross,fama1993common,hou2017replicating,kakushadze2016101}. Later approaches automated factor construction
through genetic programming, reinforcement learning, and generative
search \cite{koza1994genetic,zhaofan2022genetic,zhang2020autoalpha,cui2021alphaevolve,yu2023generating,shi2025alphaforge}. While these methods expanded the space of executable
factor constructions, their search objects remained formulas or program
structures, leaving the underlying trading hypotheses implicit before
implementation.
Recent LLM-based systems extend alpha mining by introducing planning,
memory, trajectory evolution, and guarded execution. FAMA and
AlphaAgent improve factor generation through experience reuse and
hypothesis-aware selection, while MCTS-Alpha, FactorEngine, QuantaAlpha,
Hubble, CogAlpha, and AlphaSAGE explore increasingly structured search
processes over formulas, programs, or mining trajectories
\cite{li2024can,tang2025alphaagent,shi2026navigating,lin2026factorengine,han2026quantaalpha,shi2026hubble,liu2026cognitive,chen2025alphasage}. However, these systems still largely evaluate candidates
after concrete realization, making it difficult to learn reusable priors
over trading mechanisms before implementation.
More broadly, agentic and self-evolving systems demonstrate the value of
memory, structured feedback, and evaluator-driven improvement
\cite{yao2023tree,romera2024mathematical,novikov2025alphaevolve,hu2026controlled}. Inspired by these developments, \method{} changes the search
abstraction itself by introducing explicit semantic plans over event,
context, qualities, direction, and output. This allows alpha mining to
learn and navigate a reusable semantic space of trading mechanisms before
code realization, while preserving executable validation through the
implementation stage.

\section{Conclusion}
\label{sec:conclusion}

We introduced \method{}, a framework that shifts alpha mining from
searching over factor implementations to exploring structured trading
semantics. By representing candidate factors as semantic plans before
code realization, the framework enables learning and navigating a
reusable search space of trading mechanisms. Experiments on the Chinese
stock market show that the discovered semantic factors achieve strong
predictive and portfolio performance. Further analyses demonstrate that
the semantic space contains learnable reward structures, supports
adaptive exploration under limited realization budgets, and remains
robust across different LLM realizations. These results suggest that
semantic-level search provides a promising direction for scalable and
interpretable automated alpha discovery.

\bibliography{references}
\bibliographystyle{aaai}

\appendix
\section{Experimental and Evaluation Details}
\label{sec:appendix_experimental_details}
\label{sec:appendix_eval_details}

\subsection{Dataset Split and Universe}
\label{sec:appendix_exp_setup}

We use CSI300 as the stock universe and SH000300 as the benchmark index. The time span is split into training (Jan 1, 2016 -- Dec 31, 2020), validation (Jan 1, 2021 -- Dec 31, 2022), and testing (Jan 1, 2023 -- Dec 31, 2025). The main label is the 5-day forward return:
\begin{equation}
y_{i,t}=\frac{\mathrm{Ref}(\mathrm{close},-6)}{\mathrm{Ref}(\mathrm{close},-1)}-1.
\end{equation}
Factor selection, correlation filtering, and model selection are performed only on the training and validation windows. The 2023--2025 period is reserved for held-out reporting.

\subsection{Search and Realization Configuration}
\label{sec:appendix_search_config}

The main search space contains 140 price-volume semantic components: 40
Events, 40 Contexts, 50 Qualities, 3 Directions, and 7 Outputs. These
components are defined over OHLCV and VWAP information. Additional
fundamental schema fields are enabled only in the \textsc{+Fundamental}
variant. DeepSeek-V4-Flash is used as the default realization backend due
to its low inference cost. Each run evaluates 16 schema plans per round for
80 rounds. The first 10 rounds use exploration only; later rounds gradually
introduce surrogate-guided selection and mutation.

A realization is admitted only after passing execution checks, data-contract
checks, numerical-validity checks, and look-ahead leakage screening. Plans
whose implementations still fail after one repair attempt are kept in the
reward buffer with zero reward. As in Section~\ref{sec:methodology}, each
selected plan is implemented at fast and slow time scales. For a valid
realization \(f_{p,s}\), the search reward is
\begin{equation}
\begin{aligned}
r_{p,s}
&=
\alpha\,\mathrm{RankIC}(f_{p,s})
+\beta\,\mathrm{RankICIR}(f_{p,s}) \\
&\quad
-\lambda\,\Delta_{\mathrm{lag}}(f_{p,s}),
\end{aligned}
\end{equation}
and the plan reward is
\(r(p)=\max_{s\in\{\mathrm{fast},\mathrm{slow}\}} r_{p,s}\). Unless otherwise
stated, the main experiments use \((\alpha,\beta,\lambda)=(10,1,2)\). This
reward is used for online semantic reward learning and final factor-pool
selection; IC and ICIR are reported as evaluation metrics rather than as terms
in the search reward.

\subsection{Adaptive Selector Details}
\label{sec:appendix_selector_details}

The post-cold-start selector allocates each round's implementation budget
across three sources. At each round, we randomly sample 10,000 candidate plans
from the schema space before deduplication and validity filtering. Structural
exploration favors under-covered event-context structures, while
surrogate-guided exploitation ranks sampled plans by the LightGBM predicted
score \(\hat r_t(p)\). Local mutation starts from high-reward evaluated plans
and changes one schema component at a time. These quotas are applied before
code realization, so all selected plans still pass through the same
implementation, execution-validation, leakage-screening, and reward-evaluation
pipeline.

\subsection{Exploration Novelty and Local Mutation}
\label{sec:appendix_explore_mutation}

Let \(p=(e,c,Q,d,o)\) be a sampled candidate plan, and let \(N_E(e)\),
\(N_C(c)\), and \(N_{EC}(e,c)\) denote the numbers of previously evaluated
plans containing event \(e\), context \(c\), and their pair. The exploration
score is
\[
\begin{aligned}
\nu(p)
&=
0.60\frac{1}{\sqrt{1+N_{EC}(e,c)}}
+
0.25\frac{1}{\sqrt{1+N_E(e)}} \\
&\quad
+
0.15\frac{1}{\sqrt{1+N_C(c)}} .
\end{aligned}
\]
Exploration selects high-\(\nu(p)\) candidates from the sampled pool, after
removing already evaluated plans and duplicate plan keys. This score prioritizes
rare event-context combinations while still discouraging repeated use of the
same event or context alone.

Mutation uses the top reward-ranked evaluated plans as parents and applies one
local edit per candidate. The edit is sampled from replacing a quality
\((0.22)\), replacing the context \((0.16)\), replacing the output \((0.16)\),
adding a quality \((0.14)\), dropping a quality \((0.12)\), replacing the
direction \((0.12)\), or replacing the event \((0.08)\). Quality edits respect
the \(0\)--\(3\) quality constraint, and invalid, duplicate, or previously
evaluated plans are filtered. The resulting mutation candidates are then ranked
by \(\hat r_t(p)\) under the same reward model before selection.

\subsection{Accumulated Reward Buffer}
\label{sec:appendix_accumulated_buffer}

The surrogate is trained on the accumulated reward buffer rather than a sliding
window. Each evaluated plan provides costly supervision about the relationship
between schema structure and realized reward, so retaining all observations
improves coverage of the sparse semantic space and reduces surrogate
estimation variance.

Sliding windows are useful in some reinforcement-learning settings because the
policy, state distribution, or environment may change over time, making old
samples less relevant. In our offline discovery protocol, search rounds index
the order of plan discovery under a fixed evaluation protocol. Older labels
therefore remain valid supervision, and discarding them would weaken later
reward-guided exploitation and mutation.

\subsection{Final Factor-Pool Selection}

The final modeling pool is selected from the validated factor pool by a reward-ranked filter rule. Let \(\mathcal{V}\) denote the set of valid candidate factors with available search rewards as defined in Appendix~\ref{sec:appendix_search_config}. Candidates are first sorted in descending order by this reward. We then scan this ordered list greedily and add a candidate \(f\) to the selected pool \(\mathcal{S}\) only if
\begin{equation}
\max_{g\in\mathcal{S}} |\mathrm{corr}(f,g)| < 0.7,
\end{equation}
where the correlation is computed from factor values on the pool-selection window. This rule prioritizes high-reward factors while controlling redundancy in the exported pool. The reported +Fundamental export contains the 120 selected OHLCV factors together with 30 additional factors whose schema plans use fundamental fields.

\subsection{Downstream Combiner for Factor Pools}
\label{sec:appendix_downstream_combiner}

For methods that produce an explicit factor pool, the selected factors are combined by a LightGBM ranker. Given an exported factor pool \(\mathcal{F}=\{f_1,\ldots,f_K\}\), we construct
\begin{equation}
z_{i,t}=[f_1(i,t),\ldots,f_K(i,t)]
\end{equation}
for stock \(i\) on day \(t\). Missing values and infinite values are processed before training, invalid labels are removed, and both features and labels are normalized by cross-sectional rank normalization. LightGBM is trained on the training window, while early stopping and hyperparameter selection use the validation window before test evaluation. The factor-pool experiments use 500 boosting rounds, early stopping after 50 rounds without validation improvement, and a fixed hyperparameter configuration across the corresponding factor-pool runs.

The OHLCV \method{} pool in Table~\ref{tab:main_eval} contains 120 selected
factors. The \textsc{+Fundamental} pool augments this set with 30 additional
factors whose schema plans use fundamental fields, yielding a 150-factor
pool. All learned factor-pool baselines in the NAV comparison use the same
combiner and Top50/Drop5 backtest protocol unless otherwise stated.

\subsection{Predictive Metrics}

We follow the standard factor-evaluation convention used in Qlib-style alpha-mining papers and report both linear-correlation and rank-correlation metrics. For each test day \(t\), let \(\hat y_{i,t}\) be the model score, \(y_{i,t}\) the realized label, and \(\mathcal{U}_t\) the tradable cross-section after suspension, missing-label, and validity filters.

\textbf{Information Coefficient (IC).}
IC measures whether predicted scores are linearly aligned with realized cross-sectional returns:
\begin{equation}
\begin{aligned}
\mathrm{IC}_t
&=\mathrm{corr}_{i\in\mathcal{U}_t}(\hat y_{i,t},y_{i,t}),\\
\mathrm{IC}
&=\frac{1}{|\mathcal{T}_{\mathrm{test}}|}
  \sum_{t\in\mathcal{T}_{\mathrm{test}}}\mathrm{IC}_t .
\end{aligned}
\end{equation}

\textbf{IC Information Ratio (ICIR).}
\begin{equation}
\mathrm{ICIR}=\frac{\mathrm{mean}_{t}(\mathrm{IC}_t)}{\mathrm{std}_{t}(\mathrm{IC}_t)}.
\end{equation}

\textbf{Rank Information Coefficient (Rank IC).}
\begin{equation}
\begin{aligned}
\mathrm{RankIC}_t
&=\mathrm{corr}_{i\in\mathcal{U}_t}
  \bigl(\mathrm{rank}(\hat y_{i,t}),\mathrm{rank}(y_{i,t})\bigr),\\
\mathrm{RankIC}
&=\frac{1}{|\mathcal{T}_{\mathrm{test}}|}
  \sum_{t\in\mathcal{T}_{\mathrm{test}}}\mathrm{RankIC}_t .
\end{aligned}
\end{equation}

\textbf{Rank IC Information Ratio (Rank ICIR).}
\begin{equation}
\mathrm{RankICIR}=\frac{\mathrm{mean}_{t}(\mathrm{RankIC}_t)}{\mathrm{std}_{t}(\mathrm{RankIC}_t)}.
\end{equation}
Predictive metrics are computed on the held-out test/backtest window and do not include transaction costs. Days with too few tradable assets or degenerate prediction/label vectors are excluded.

\subsection{Trading Strategy}

Strategy-level evaluation uses Qlib's \texttt{TopkDropoutStrategy} with \texttt{topk=50} and \texttt{n\_drop=5}. On each rebalance date, stocks are ranked by \(\hat y_{i,t}\). The target portfolio holds the top 50 names with equal weights. To reduce turnover, existing holdings that remain sufficiently highly ranked are retained; at most five lowest-ranked current holdings are replaced by the highest-ranked non-held names. Rebalancing is daily. This Top50/Drop5 rule follows recent Qlib-based alpha-mining protocols and keeps the comparison focused on signal quality rather than portfolio optimization.

\subsection{Execution Assumptions and Costs}

Trades are executed at the \textbf{open price} on the trading day after signal generation (\texttt{deal\_price=open}). Transaction costs are modeled as a buying cost of 0.05\% and a selling cost of 0.15\%, with a minimum fee of 5 currency units per trade and a round-trip friction of approximately 0.20\%. Limit-up/limit-down filtering is enabled with \texttt{limit\_threshold=0.095}. Untradable names are skipped for new purchases, while existing holdings are handled according to Qlib exchange rules.

\subsection{Portfolio Metrics}

Predictive metrics measure ranking quality but not tradability. We therefore report portfolio-level metrics under the Top50/Drop5 strategy. Let \(g^{\mathrm{port}}_t\) be the gross portfolio return before transaction costs, \(g^{\mathrm{bench}}_t\) the CSI300 benchmark return, and \(c_t\) the transaction-cost term deducted once by the backtest. The daily excess return is defined as
\begin{equation}
e_t = g^{\mathrm{port}}_t - g^{\mathrm{bench}}_t - c_t,
\end{equation}
where the benchmark is SH000300. All portfolio metrics below are computed from \(\{e_t\}\), with \(N\) denoting the number of trading days in the evaluation window and 252 trading days per year.

\textbf{Annualized Excess Return (AER).}
\begin{equation}
\mathrm{AER}=\left(\prod_{t=1}^{N}(1+e_t)\right)^{252/N}-1.
\end{equation}
AER is therefore the compound annualized benchmark-relative excess return
after transaction costs, rather than the annualized raw portfolio return.

\textbf{Information Ratio (IR).}
\begin{equation}
\mathrm{IR}=\frac{\mathrm{mean}_t(e_t)}{\mathrm{std}_t(e_t)}\sqrt{252}.
\end{equation}
Because the excess return is defined relative to CSI300 rather than a risk-free rate, IR measures benchmark-relative risk-adjusted performance.

\textbf{Maximum Drawdown (MDD).}
MDD is computed on the same cumulative excess-return curve induced by
\(\{e_t\}\):
\begin{equation}
\mathrm{MDD}=\max_{t\leq N}\left(1-\frac{C_t}{\max_{s\leq t}C_s}\right),
\quad
C_t=\prod_{s=1}^{t}(1+e_s).
\end{equation}

\section{Additional Market Results}
\label{sec:appendix_csi500}

\begin{figure*}[t]
    \centering
    \includegraphics[width=0.86\textwidth]{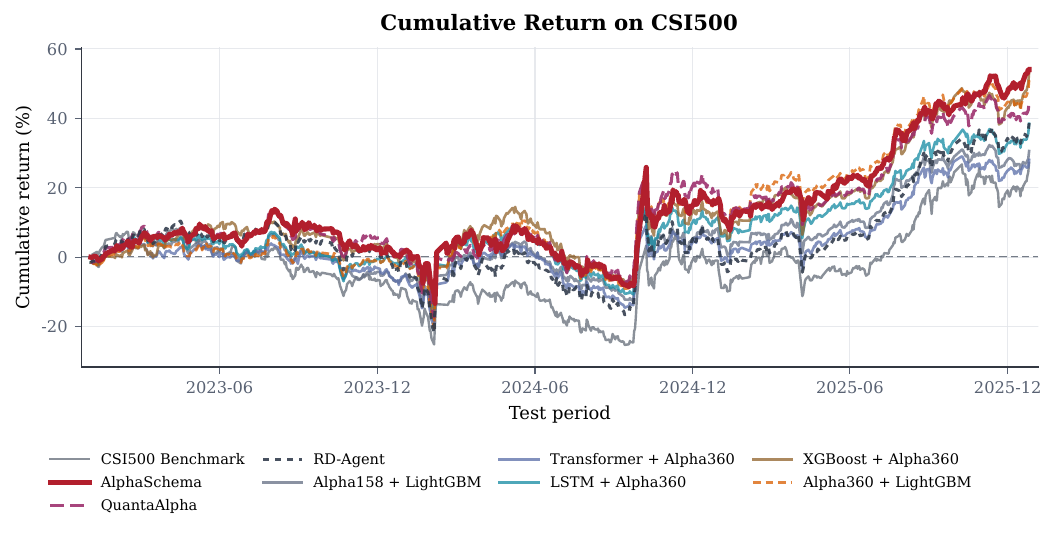}
    \caption{CSI500 held-out cumulative-return curves over the 2023--2025 test period. The factor pool is independently mined and selected on CSI500 using a configuration analogous to CSI300; the LightGBM combiner is trained and validated on CSI500 with the same downstream protocol.}
    \label{fig:appendix_csi500_cumulative_return}
\end{figure*}

We include CSI500 as an additional held-out universe to test whether the same
semantic-mining protocol remains effective beyond the CSI300 setting used in
the main paper. The protocol mirrors Appendix~\ref{sec:appendix_eval_details}:
the training, validation, and testing windows remain unchanged, the benchmark
is replaced by the CSI500 index, and the downstream Top50/Drop5 strategy is
kept fixed. Rather than transferring the CSI300 factor pool, we conduct an
independent multi-round discovery run on CSI500 with the same reward
definition and a configuration analogous to the CSI300 setting. Candidate
factors are then sorted directly by the CSI500 reward and filtered by the same
correlation-control rule to form the CSI500 factor pool. The LightGBM combiner
is trained on the CSI500 training window and selected on the CSI500 validation
window. It uses the same preprocessing, feature normalization, hyperparameters,
early-stopping rule, and backtest configuration as the CSI300 protocol.
Figure~\ref{fig:appendix_csi500_cumulative_return}
reports the resulting cumulative-return curves over the 2023--2025 test
period. The \method{} curve remains in the upper envelope of the comparison
set, providing an out-of-universe check of the full discovery-and-selection
pipeline rather than a transfer evaluation of a CSI300-mined factor pool.

\subsection{Factor Decay Analysis}
\label{sec:appendix_factor_decay}

We analyze factor decay on daily Rank IC observations restricted to the
held-out 2023--2025 test period. For each test trading day, we compute
cross-sectional Rank IC between each factor and the 5-day open-to-open
executable return label, apply the fixed factor-direction vector supplied with
the evaluation artifact, and average the direction-adjusted Rank IC values
across each pool. We then compute a 252-trading-day rolling mean using only
these test-period daily observations. Consequently, the curve begins after the
first complete 252-day test window, on Jan 15, 2024. The direction convention
is held fixed to match the supplied evaluation artifact. This diagnostic uses
the executable open-to-open label available in the factor-decay artifact; the
main predictive metrics in
Table~\ref{tab:main_eval} use the close-to-close label defined in
Appendix~\ref{sec:appendix_exp_setup}.

\begin{figure}[!t]
    \centering
    \includegraphics[width=\columnwidth]{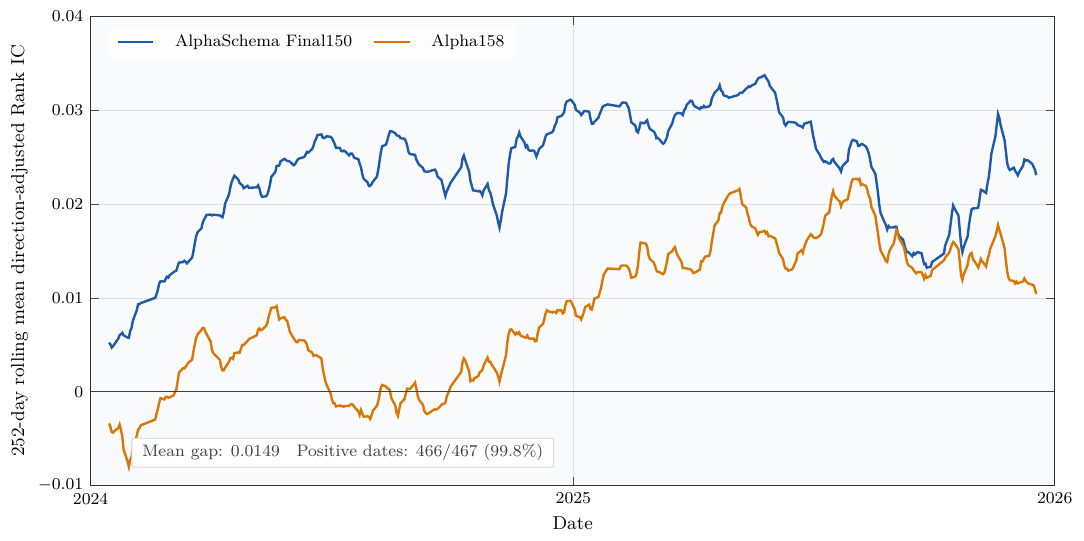}
    \caption{Test-period factor-decay analysis. We plot the test-only 252-day
    rolling mean direction-adjusted Rank IC of the exported 150-factor pool
    and Alpha158 from Jan 2024 to Dec 2025, using a 5-day open-to-open
    executable return label.}
    \label{fig:appendix_factor_decay}
\end{figure}

Figure~\ref{fig:appendix_factor_decay} shows that the exported pool remains
above Alpha158 on 466 of 467 valid test-period rolling dates. The mean rolling
Rank IC gap is 0.0149. The year-end gaps are 0.0216 in 2024 and 0.0127 in
2025. This test-period view indicates that the final factor pool retains more
persistent ranking signal than the Alpha158 reference library over the test
period.

\section{Construction of the Schema Vocabulary}
\label{sec:appendix_schema_construction}

We construct a practical, extensible vocabulary rather than an exhaustive
ontology of market mechanisms. Each record has a clear functional role and can
be recombined, implemented, and evaluated as part of a schema plan.

\subsection{Formal Schema Notation}
\label{sec:appendix_schema_notation}

The semantic plan space is written as
\[
  \mathcal{P} = \mathcal{V}_{E} \times \mathcal{V}_{C} \times \mathcal{Q}_{\mathrm{set}}
                \times \mathcal{V}_{D} \times \mathcal{V}_{O}.
\]
Here \(\mathcal{P}\) denotes the set of all candidate plans. The five component
sets correspond to \emph{Event} choices \(\mathcal{V}_{E}\), \emph{Context} choices
\(\mathcal{V}_{C}\), \emph{Quality} choices \(\mathcal{V}_{Q}\), \emph{Direction}
choices \(\mathcal{V}_{D}\), and \emph{Output} choices \(\mathcal{V}_{O}\). The
quality-set choices are
\[
  \mathcal{Q}_{\mathrm{set}}
  =
  \{Q\subseteq\mathcal{V}_{Q}:0\leq |Q|\leq 3\},
\]
so a plan may include zero to three confirmation or filtering conditions,
whereas the other fields usually select one primary option.

Thus, a plan can be written as \(p=(e,c,Q,d,o)\), where \(e\in\mathcal{V}_{E}\),
\(c\in\mathcal{V}_{C}\), \(Q\in\mathcal{Q}_{\mathrm{set}}\),
\(d\in\mathcal{V}_{D}\), and \(o\in\mathcal{V}_{O}\). Its canonical key is
\[
  \operatorname{key}(p)=e\mid c\mid q_1\mid\cdots\mid q_m\mid d\mid o,
\]
which supports deduplication, coverage tracking, mutation, and surrogate
scoring.

\subsection{Functional Roles and Compositional Grammar}

Each category has a designated role. An \emph{Event} identifies what happened;
a \emph{Context} specifies the market state or reference condition;
\emph{Qualities} confirm, filter, or reweight the event; \emph{Direction}
relates it to future returns; and \emph{Output} specifies the numerical form of
the tradable signal. Together they describe what happened, under which
conditions, with what supporting evidence, how it should be traded, and how it
should be expressed. A plan is represented as

\begin{equation}
    p=(e,c,Q,d,o),
\end{equation}

where \(Q\) may contain zero to three complementary quality conditions. The plan is
therefore compositional and searchable, but leaves its mathematical realization
to the implementation agent.

These roles are functional rather than strictly independent. Event, Context,
and Quality may overlap; for example, volume expansion can be an event or a
confirmation of a price event. This helps explain the ablation results in
Section~\ref{sec:appendix_ablation}: the agent can recover related cues from
the remaining fields and often retain executable code, but cannot fully recover
the missing semantic constraint, so factor quality declines more than validity.

\subsection{JSON Representation and Semantic Constraints}

Each vocabulary element is a structured JSON record whose core fields are
summarized in Table~\ref{tab:appendix_schema_json_fields}.

\begin{table}[t]
\centering
\caption{Core fields of an individual schema record.}
\label{tab:appendix_schema_json_fields}
\small
\begin{tabularx}{\columnwidth}{lX}
\toprule
\textbf{Field} & \textbf{Role} \\
\midrule
\texttt{schema\_id} & Stable identifier used in plan construction, logging,
and deduplication. \\
\texttt{category} & Functional class: Event, Context, Quality, Direction, or
Output. \\
\texttt{scope} & Whether the concept is defined time-series-wise or
cross-sectionally. \\
\texttt{name} & Short human-readable name of the market concept. \\
\texttt{description} & Primary statement of the intended market logic. \\
\texttt{terms} & Supplementary definitions that disambiguate important terms
appearing in the description. \\
\bottomrule
\end{tabularx}
\end{table}

The \texttt{description} carries the market logic, and \texttt{terms}
disambiguates its key concepts; the remaining fields support identification and
composition. No field prescribes an operator, function, window, threshold, or
parameter. The code agent chooses these under the data contract and execution
constraints. Event, Context, and Quality also avoid unnecessarily fixing a
continuation or reversal interpretation, which is primarily assigned to
Direction. The same plan can thus admit multiple implementations without
changing its semantics.

\subsection{Sources, Abstraction, and Iterative Expansion}

Sources include sell-side reports, papers, books, practitioner blogs, and
researcher objectives, such as dedicated coverage of futures term structure,
inventory, or warehouse receipts. We extract the underlying market mechanism,
remove implementation-specific formulas and parameters, assign its category
and scope, define its key terms, and check it for duplication and coverage.

Between discovery cycles, the vocabulary grows in two ways: semantic mutations
of records associated with strong plans deepen promising regions, while new
mechanisms absent from the existing schema and factor pools expand coverage in
approximately orthogonal directions. This offline vocabulary expansion creates
or revises records and is distinct from online plan mutation, which only
recombines records already in the vocabulary.

\section{Example Schema-to-Factor Realization}
\label{sec:appendix_alpha_code}

This section gives one concrete example of how a semantic schema plan is
realized into executable factor code under the CSI300 stock-data contract. The
factor is generated by DeepSeek-V4-Flash. Using 5-day open-to-open future
returns as labels, its best period is 20 and its Rank IC on CSI300 is 0.0462 on
2016--2021 training, 0.0483 on 2022 validation, and 0.0474 on 2023--2025 test.

\begin{table}[t]
\centering
\caption{CSI300 Rank IC of the example schema-realized factor using 5-day
open-to-open future returns as labels.}
\label{tab:appendix_example_factor_rankic}
\small
\begin{tabular}{lcc}
\toprule
\textbf{Split} & \textbf{Window} & \textbf{Rank IC} \\
\midrule
Train & 2016--2021 & 0.0462 \\
Validation & 2022 & 0.0483 \\
Test & 2023--2025 & 0.0474 \\
\bottomrule
\end{tabular}
\end{table}

The plan measures whether price movement and trading-volume change are jointly
strong relative to the same-day universe, conditions this event on the stock's
distance from its own VWAP or moving-average reference, downweights repeated
events through the Quality component, keeps a continuation interpretation, and
stores the result through causal event memory.

The corresponding user message specifies the semantic composition order, the
implementation workflow, cross-sectional leakage constraints, and the complete
task JSON. The concrete JSON plan supplied for this example is reported next.

\subsection{Concrete Schema JSON Records}

The following records list one complete JSON object from each schema class in
the same plan.

\noindent\textbf{Event schema.}
\begin{lstlisting}[style=appendixlisting]
{
  "schema_id": "event.cross_section_effort_result_breakout_rank",
  "category": "event",
  "scope": "cross_sectional",
  "name": "Cross-sectional effort-result breakout rank",
  "description": "Compares stocks by the efficiency strength with which input-variable changes and price outcomes jointly form a directional breakout.",
  "terms": {
    "effort-result breakout": "input variables such as trading volume change while price also shows a directional movement.",
    "efficiency strength": "the relative degree of synchronization and amplification between input changes and price outcomes."
  }
}
\end{lstlisting}

\noindent\textbf{Context schema.}
\begin{lstlisting}[style=appendixlisting]
{
  "schema_id": "context.cross_section_reference_distance_rank",
  "category": "context",
  "scope": "cross_sectional",
  "name": "Cross-sectional reference-distance rank",
  "description": "Compares stocks by the standardized distance between the event location and each stock's own reference level, such as VWAP or a moving average.",
  "terms": {
    "reference distance": "the standardized distance from pivot, channel, VWAP, moving-average, or other stock-specific reference locations.",
    "relative position context": "same-date comparison of distance or proximity across stocks."
  }
}
\end{lstlisting}

\noindent\textbf{Quality schema.}
\begin{lstlisting}[style=appendixlisting]
{
  "schema_id": "quality.event_cooldown",
  "category": "quality",
  "scope": "time_series",
  "quality_type": "filter",
  "name": "Event-cooldown filter",
  "description": "Downweight or filter same-type events when they occur repeatedly in a short period, reducing repeated-event stacking.",
  "terms": {
    "event cooldown": "a spacing or downweighting rule between similar events.",
    "repeated-event stacking": "multiple similar events appearing within a short interval."
  }
}
\end{lstlisting}

\noindent\textbf{Direction schema.}
\begin{lstlisting}[style=appendixlisting]
{
  "schema_id": "direction.continuation",
  "category": "direction",
  "scope": "time_series",
  "name": "Continuation direction guidance",
  "description": "Treat the event as an alpha-construction tendency in which price is expected to continue along the event-implied direction.",
  "terms": {
    "continuation": "after the event occurs, price continues along the event-implied direction.",
    "event-implied direction": "a direction obtained from breakouts, returns, deviations, or relative strength."
  }
}
\end{lstlisting}

\noindent\textbf{Output schema.}
\begin{lstlisting}[style=appendixlisting]
{
  "schema_id": "output.event_decay_signal",
  "category": "output",
  "scope": "time_series",
  "name": "Event-decay signal output",
  "description": "Keeps finite memory after a valid event trigger and gradually decays the final signal as the event becomes older.",
  "terms": {
    "finite memory": "the event affects the signal only within a limited observation window.",
    "event age": "the time elapsed since the most recent event trigger or confirmation."
  }
}
\end{lstlisting}

\subsection{Schema-to-Code Mapping and Complete Factor}

The complete implementation below is a stock-compatible realization of the
schema records above. It makes the causal data handling and the five
schema-to-code correspondences explicit.

\noindent\textbf{Complete realized alpha factor.}
\begin{lstlisting}[style=appendixlisting,language=Python]
# Implementation plan:
# event: compute price-volume synergy and cross-sectional z-score it.
# context: z-score the distance from VWAP or SMA across stocks.
# quality: cool repeated events using a causal EWM-based downweighting term.
# direction: keep the sign of the original price-volume synergy.
# output: store the signal through causal exponential event memory.

import numpy as np
import pandas as pd

FACTOR_META = {
    "source_plan_id": (
        "event.cross_section_effort_result_breakout_rank|"
        "context.cross_section_reference_distance_rank|"
        "quality.event_cooldown|direction.continuation|"
        "output.event_decay_signal"
    ),
    "name": "effort_result_reference_cooldown_memory",
    "input_type": "cross_sectional_panel",
    "periods": [20, 100],
    "best_period": 20,
    "period_reason": (
        "best period=20 for short-term effort-result behavior; "
        "period=100 provides a longer event-memory and cooldown reference"
    ),
    "uses_columns": ["close", "volume", "vwap"]
}

def F_some_factor_name(
    panel: dict[str, pd.DataFrame],
    period: int
) -> pd.DataFrame:
    close = pd.DataFrame({
        s: df["close"].astype(float) for s, df in panel.items()
    }).sort_index()

    volume = pd.DataFrame({
        s: df["volume"].astype(float)
        for s, df in panel.items() if "volume" in df
    }).sort_index()
    if volume.empty:
        amount = pd.DataFrame({
            s: df["amount"].astype(float)
            for s, df in panel.items() if "amount" in df
        }).sort_index()
        if not amount.empty:
            volume = amount / close.replace(0, np.nan) * 100
        else:
            volume = pd.DataFrame(1.0, index=close.index, columns=close.columns)

    common_idx = close.index.intersection(volume.index)
    close = close.loc[common_idx]
    volume = volume.loc[common_idx].fillna(0)

    pct_chg = close.pct_change().fillna(0)
    vol_chg = volume.pct_change().fillna(0)
    synergy = pct_chg * vol_chg

    x_mean = synergy.mean(axis=1, skipna=True)
    x_std = synergy.std(axis=1, skipna=True).replace(0, 1)
    event_z = synergy.sub(x_mean, axis=0).div(x_std, axis=0).fillna(0)

    try:
        first_sym = next(iter(panel))
        if "vwap" in panel[first_sym].columns:
            vwap = pd.DataFrame({
                s: df["vwap"].astype(float) for s, df in panel.items()
            }).sort_index().loc[common_idx]
        else:
            vwap = None
    except (StopIteration, KeyError):
        vwap = None

    if vwap is not None:
        distance = (close - vwap).abs() / vwap.replace(0, np.nan)
    else:
        sma = close.rolling(period, min_periods=1).mean()
        distance = (close - sma).abs() / sma.replace(0, np.nan)
    distance = distance.fillna(0)

    d_mean = distance.mean(axis=1, skipna=True)
    d_std = distance.std(axis=1, skipna=True).replace(0, 1)
    context_z = distance.sub(d_mean, axis=0).div(d_std, axis=0).fillna(0)

    combined = event_z + 0.5 * context_z
    ewm_combined = combined.ewm(span=period, adjust=False).mean()
    cooling_factor = 1.0 / (1.0 + ewm_combined.abs() * 0.3)
    cooled = combined * cooling_factor

    direction = np.sign(pct_chg * vol_chg).replace(0, 0.01)
    interim = cooled.abs() * direction

    signal = interim.ewm(span=period, adjust=False).mean()
    signal = signal.fillna(0)
    signal = signal.replace([np.inf, -np.inf], 0)
    signal = signal.clip(-5, 5)

    return signal
\end{lstlisting}

This example illustrates why the schema is useful as an intermediate search
object. The generated code is not merely an opaque formula: each block has a
direct correspondence to the event, context, quality, direction, and output
fields. Before inclusion in the final pool, such implementations are subjected
to execution validation, numerical checks, look-ahead screening, reward
evaluation, and correlation-aware pool selection.

\subsection{Implementation-Agent System Prompt}
\label{sec:appendix_code_agent_prompt}

The system instruction associated with the stock cross-sectional realization
example is reported below. Tasks containing a schema with
\texttt{scope="cross\_sectional"} use this cross-sectional prompt variant. The
selected semantic plan is supplied separately in the user message as a
structured JSON object.

\begin{lstlisting}[style=appendixlisting]
You are a stock cross-sectional factor implementation agent. Given a
factor plan, write a vectorized pandas factor function.

The task contains at least one schema with scope="cross_sectional", so
the implementation may combine causal time-series features with
same-date cross-sectional comparison, normalization, or neutralization.

Hard requirements:
- Output only one Python file, including required imports.
- Begin with concise comments mapping the implementation to event,
  context, quality, direction, and output.
- Use the signature
  `def F_some_factor_name(panel: dict[str, pd.DataFrame],
      period: int) -> pd.DataFrame:`.
- `panel` maps each stock symbol to a daily OHLCV/VWAP DataFrame.
- Return a date-by-symbol `pd.DataFrame` factor matrix.
- Use `period` as the only primary parameter; derive secondary windows,
  thresholds, and smoothers from it.
- Include `FACTOR_META` with input type, name, source plan id, periods,
  period reason, and used columns.
- The signal at time t may only use information available at or before t.
- Same-date cross-sectional ranking, z-scoring, winsorization,
  demeaning, and neutralization are allowed when required by the schema.
- Prefer vectorized pandas operations such as `shift`, `rolling`,
  `ewm`, `rank(axis=1)`, `where`, `clip`, and `tanh`.
- Return finite values and degrade gracefully if optional columns are
  unavailable.
\end{lstlisting}

\section{Analysis Experiment Details}
\label{sec:appendix_ablation}

\subsection{Semantic Navigation Details}
\label{sec:appendix_semantic_navigation}

The semantic-navigation analysis maps the structured plans selected during one
complete multibar search run into a continuous embedding space. Each plan is
converted to a canonical text containing its Event, Context, Qualities,
Direction, and Output descriptions; reward, round index, selection policy,
validation status, and generated code are excluded from the embedded text. We
then L2-normalize the 1024-dimensional text embeddings, reduce them to 50 PCA
components, and run K-means with 12 clusters. The first two PCA coordinates are
used only for visualization, while the cluster assignments are obtained in the
50-dimensional PCA space.

For each round \(t\) and manifold \(k\), we compute
\[
  p_t(k)=
  \frac{\#\{\text{selected plans in manifold }k\text{ at round }t\}}
       {\#\{\text{selected plans at round }t\}} .
\]
The dominant manifold of a round is the region with largest \(p_t(k)\). To
interpret these regions, we reverse-map each dominant manifold to the most
frequent schema components among plans assigned to that cluster.
Table~\ref{tab:appendix_dominant_manifold_semantics} reports the corresponding
semantic schema patterns. This table complements the trajectory visualization
in the main text by making the dominant regions economically interpretable.

\begin{table*}[!t]
\centering
\caption{Semantic interpretation of the dominant manifolds along the search
trajectory. Each dominant schema is obtained by reverse-mapping the manifold to
the most frequent Event, Context, and Output components in that cluster.}
\label{tab:appendix_dominant_manifold_semantics}
\setlength\tabcolsep{3.5pt}
\scriptsize
\resizebox{\textwidth}{!}{%
\begin{tabular}{lccclcc}
\toprule
\textbf{Stage} & \textbf{Rounds} & \textbf{Cluster} & \textbf{Mode} &
\textbf{Dominant semantic schema} & \textbf{Mean reward} & \textbf{Mass} \\
\midrule
M1 & 0--3 & 8 & Explore &
entropy compression then directional move; multi-session reference; persistent condition signal
& 1.415 & 0.239 \\
M2 & 4--28 & 3 & Explore &
close-position extreme; multi-window volatility conflict; event-decay signal
& 1.540 & 0.276 \\
M3 & 29--36 & 1 & Explore &
effort-result breakout; multi-session reference; bounded continuous signal
& 1.626 & 0.283 \\
M4 & 37--42 & 9 & Mutation &
short-long efficiency conflict; multi-session reference; persistent condition signal
& 2.039 & 0.240 \\
M5 & 43--48 & 1 & Explore &
effort-result breakout; multi-session reference; bounded continuous signal
& 1.626 & 0.185 \\
M6 & 49--72 & 11 & Mutation &
short-long efficiency conflict; multi-session reference; event-decay signal
& 2.236 & 0.259 \\
M7 & 73--79 & 6 & Exploit &
multi-window range-efficiency alignment; close-extreme memory zone; bounded continuous signal
& 2.516 & 0.258 \\
M8 & 80--99 & 2 & Exploit &
multi-window close-extreme alignment; high-information entropy state; event-decay signal
& 2.480 & 0.297 \\
\bottomrule
\end{tabular}
}
\end{table*}

The dominant trajectory is therefore semantically interpretable. Early stages
mostly occupy broad exploratory regions related to compression, local extremes,
and breakout-style events. Later stages shift toward reward-enriched regions
defined by short--long efficiency conflicts and multi-window alignment
patterns. The final two dominant manifolds are especially concentrated:
cluster 6 is dominated by multi-window range-efficiency alignment with bounded
continuous outputs, while cluster 2 is dominated by multi-window close-extreme
alignment with event-decay outputs. This indicates that the search does not
merely move through an embedding space; it reallocates evaluations toward
specific schema combinations with higher average realized reward.

\subsection{Realization-Budget Efficiency Details}
\label{sec:appendix_realization_budget_efficiency}

The realization-budget experiment complements the main search results by
measuring how many independent code realizations should be spent on the same
semantic plan. We use 100 schema plans with up to eight realizations per plan.
Each realization shares the same semantic plan and implementation prompt but
uses an independent generation attempt and one repair opportunity. To avoid
mixing realization variance with period-selection variance, all successful
realizations are re-materialized and re-evaluated at a fixed period of 30.

For plan \(i\), let \(a_{i,j}\) denote the absolute Rank IC of its \(j\)-th
successful realization. We define an approximate reference quality by averaging
all available successful realizations,
\[
  q_i^{\mathrm{RIC}} = \frac{1}{m_i}\sum_{j=1}^{m_i} a_{i,j},
\]
where \(m_i\) is the number of successful realizations for that plan. For each
budget \(k\in\{1,\ldots,5\}\), we repeatedly sample \(k\) realizations per plan,
average them to estimate plan quality, and compare the induced plan ranking
with the ranking from \(q_i^{\mathrm{RIC}}\). The bootstrap uses 3,000 repetitions and reports
Spearman correlation, mean absolute error, top-20\% recall, and recall per
realization.

\begin{table}[!t]
\centering
\caption{Realization-budget efficiency under repeated implementations of the
same semantic plans. Increasing \(k\) improves plan-quality estimation, but the
top-plan recall obtained per realization decreases.}
\label{tab:appendix_realization_budget_efficiency}
\setlength\tabcolsep{3.5pt}
\small
\resizebox{\columnwidth}{!}{%
\begin{tabular}{ccccc}
\toprule
\(\mathbf{k}\) & \(\rho_S\) w.r.t. reference & \textbf{MAE} &
\textbf{Top-20\% Recall} & \textbf{Recall / Realization} \\
\midrule
1 & 0.445 & 0.00664 & 0.462 & 0.462 \\
2 & 0.640 & 0.00427 & 0.570 & 0.285 \\
3 & 0.763 & 0.00307 & 0.653 & 0.218 \\
4 & 0.852 & 0.00221 & 0.712 & 0.178 \\
5 & 0.919 & 0.00147 & 0.780 & 0.156 \\
\bottomrule
\end{tabular}
}
\end{table}

The results show the expected bias--variance trade-off in code realization:
more realizations give a cleaner estimate of a fixed plan's expected quality,
but each extra realization consumes budget that could have evaluated a new
semantic candidate. Since the search objective is to discover and rank many
promising plans under a fixed mining budget, the single-realization protocol is
a budget-efficient choice even though it is not the lowest-variance estimator
of any individual plan's expected reward. The reference \(q_i^{\mathrm{RIC}}\) is a finite
rollout average rather than a noiseless oracle, so this experiment should be
interpreted as an internal budget-efficiency diagnostic.

\subsection{Schema Grammar Ablation}

The schema ablation is a leave-one-component-out experiment over the five semantic fields \(E, C, Q, D, O\). We sample 100 high-reward full schema plans and construct five blind variants for each plan by removing one field at a time. The implementation agent receives only the visible fields and is not told which component was removed. This yields 500 implementation tasks, of which 466 pass the full validity and evaluation pipeline. The primary metric is mean absolute Rank IC, because this experiment measures whether a generated factor contains ranking information regardless of whether the profitable direction is positive or negative.

\begin{table}[t]
\centering
\caption{Detailed schema-grammar leave-one-out results. Full is the source full-plan reference used to construct the ablation tasks.}
\label{tab:appendix_schema_ablation_design}
\setlength\tabcolsep{3pt}
\small
\resizebox{\columnwidth}{!}{%
\begin{tabular}{lccccc}
\toprule
\textbf{Variant} & \textbf{Missing Field} & \textbf{Valid} & \(\mathbf{|RankIC|}\uparrow\) & \textbf{Rel. (\%)} & \textbf{Reward} \\
\midrule
\method{} (Full source) & -- & 100/100 & 0.0185 & 100.0 & 4.174 \\
w/o Event & Event & 93/100 & 0.0147 & 78.9 & 2.199 \\
w/o Context & Context & 94/100 & 0.0131 & 69.4 & 2.249 \\
w/o Qualities & Qualities & 95/100 & 0.0134 & 72.9 & 2.483 \\
w/o Direction & Direction & 88/100 & 0.0134 & 71.7 & 2.219 \\
w/o Output & Output & 96/100 & 0.0141 & 75.1 & 2.310 \\
\bottomrule
\end{tabular}
}
\end{table}

The full reference is not reimplemented under the blind ablation prompt; it is read from the source full-plan runs used to build the 100-plan task set. We therefore use this experiment as a component-level diagnostic rather than as a final portfolio-level comparison. A complete paired subset in which all five ablated variants are valid contains 71 source plans and yields the same qualitative ordering: every leave-one-out variant remains below the full source reference.

Figure~\ref{fig:schema_leave_one_out} visualizes the same comparison in the main paper. Every field removal reduces retained signal strength, while most variants still maintain high implementation validity. The degradation is therefore not only an execution failure effect, but also reflects weaker schema semantics.

\subsection{Why Ablated Schemas Remain Executable}
\label{sec:appendix_ablation_executable}

The high validity rate in the leave-one-component-out experiment does not
mean that the implementation agent reconstructs the omitted schema field.
Removing a schema component removes one trading-design constraint, but it
does not remove the implementation contract: the agent still receives the
stock-panel input format, admissible OHLCV/VWAP fields, causal-computation
rules, finite-output requirements, and the required date-by-symbol output
matrix. These constraints make the task operationally complete even when one
semantic field is absent.

The remaining fields also provide correlated cues. Events may imply signed
price movement, contexts may specify relative regimes, qualities may suggest
confirmation or filtering, and outputs may imply ranking, smoothing, or
bounded transformations. The agent can further close underspecified programs
with generic factor-programming priors such as returns, VWAP deviations,
rolling statistics, cross-sectional ranks or z-scores, clipping, and
\(\tanh\) compression. This is functional completion rather than recovery of
the missing design variable.

Thus, the ablation separates operational completeness from semantic
completeness. Operational completeness is largely enforced by the
implementation contract, explaining why 466 of 500 ablated tasks remain
valid. Semantic completeness depends on the five schema fields, explaining
why all leave-one-out variants still lose predictive strength and retain only
69.4\%--78.9\% of the full-source absolute Rank IC.

\subsection{Schema-Space Reward Predictability}
\label{sec:appendix_schema_predictability}

The predictability experiment evaluates whether reward can be inferred from the discrete schema representation alone. Its dataset is an aggregate reward archive collected for this analysis, rather than only the five CSI300 production search runs reported in Section~\ref{sec:experiment}. The archive contains 12,126 valid plan--reward pairs and 11,295 unique schema keys, including auxiliary search trajectories used exclusively for reward-prediction analysis. We use an 80/20 split grouped by canonical schema key, so repeated observations of the same semantic combination do not cross the train/test boundary; this yields 2,430 held-out test examples. The target is the realized scalar plan reward defined in Appendix~\ref{sec:appendix_search_config}.
The feature construction intentionally excludes schema text embeddings, TF--IDF features, generated code, factor values, and backtest trajectories. The two main feature sets are: (i) single-schema indicators with category/scope counts; and (ii) schema-pair features with pairwise schema, category, and scope interactions. The mean baseline predicts a constant reward, while the shuffled-reward control preserves the feature matrix and model class but randomly permutes reward labels.

\begin{table}[t]
\centering
\caption{Full predictability results for schema-space reward prediction. Top10\% lift is measured against random selection.}
\label{tab:appendix_schema_predictability_full}
\setlength\tabcolsep{3pt}
\small
\resizebox{\columnwidth}{!}{%
\begin{tabular}{llcccc}
\toprule
\textbf{Feature Set} & \textbf{Model} & \(\rho_S\) & \(\rho_P\) & \textbf{RMSE$\downarrow$} & \textbf{Top10\% Lift} \\
\midrule
Mean baseline & Mean & 0.000 & 0.000 & 0.0908 & 1.000 \\
Single schema & Ridge & 0.212 & 0.214 & 0.0892 & 1.413 \\
Single schema & ElasticNet & 0.214 & 0.217 & 0.0888 & 1.381 \\
Single schema & LGBM & 0.237 & 0.245 & 0.0884 & 1.506 \\
Single schema & MLP & 0.121 & 0.123 & 0.0963 & 1.295 \\
Schema + pair interaction & LGBM & \textbf{0.239} & \textbf{0.250} & \textbf{0.0884} & \textbf{1.519} \\
Schema + pair interaction & MLP & 0.204 & 0.201 & 0.0903 & 1.273 \\
Shuffled reward & LGBM & 0.029 & 0.021 & 0.0917 & 0.946 \\
\bottomrule
\end{tabular}
}
\end{table}

The shuffled-reward control remains near random, supporting that the observed predictability reflects a genuine schema--reward relationship rather than feature dimensionality alone.

\begin{figure}[t]
    \centering
    \includegraphics[width=\columnwidth]{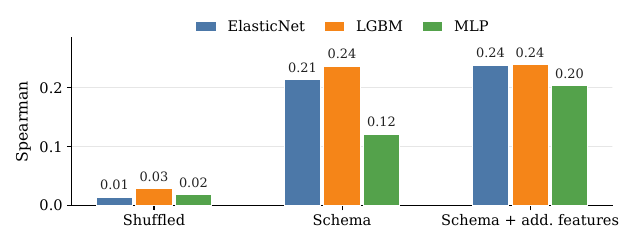}
    \caption{Reward predictability from structured schema features.}
    \label{fig:schema_predictability}
\end{figure}

Figure~\ref{fig:schema_predictability} visualizes the same control logic:
schema features yield nontrivial ranking and top-decile enrichment, pair
interactions add a small gain, and shuffled rewards collapse toward random
selection.

\subsection{Why Single-Realization Rewards Remain Predictable}
\label{sec:appendix_single_realization_predictability}

Each reward-buffer entry is a stochastic realization outcome rather than a
noiseless estimate of plan quality. Let \(r^{(k)}(p)\) denote the reward from
the \(k\)-th realization of plan \(p\), under the fixed backend, prompt,
validation checks, fast/slow realization protocol, and reward definition. Here
\(r^{(k)}(p)\) follows the same search-reward definition as \(r(p)\) in
Section~\ref{sec:methodology}, with the superscript indexing independent
realization attempts. We
write the corresponding expected realized reward as
\[
q(p)=\mathbb{E}[r^{(k)}(p)\mid p],
\]
so a single observed reward can be decomposed as
\[
r^{(k)}(p)=q(p)+\varepsilon^{(k)}(p),
\qquad
\mathbb{E}[\varepsilon^{(k)}(p)\mid p]=0.
\]
Here \(\varepsilon^{(k)}(p)\) includes implementation choices, repair paths,
and finite-sample backtest noise. The quantity \(q(p)\) is not a separate
objective; it is the expected version of the realized reward already used by
the search loop.

Single-realization labels increase estimation variance, but they do not
change the population target of schema-level reward learning. For a predictor
trained from structured features \(\phi(p)\), the squared-loss optimum is
\[
g^*(\phi)=\mathbb{E}[r^{(k)}(p)\mid \phi(p)=\phi].
\]
Because \(\phi(p)\) is a function of \(p\), the zero-mean realization noise
also satisfies
\(\mathbb{E}[\varepsilon^{(k)}(p)\mid \phi(p)]=0\) by iterated
expectations. Therefore
\[
g^*(\phi)=\mathbb{E}[q(p)\mid \phi(p)=\phi].
\]
The surrogate is thus learning the conditional average reward associated with
schema structure, not predicting the exact reward of the next random
implementation.

This averaging occurs across structurally related plans rather than through
many repetitions of the same plan. The same schema IDs, categories, scopes, and
pairwise interactions recur across many different canonical plans, allowing the
model to pool noisy observations at the feature level.
The held-out Spearman correlation and top-decile lift should therefore be
interpreted as evidence of reusable schema-level ordering information under
the historical realization and evaluation pipeline. The canonical-key group
split described above prevents exact-plan repetition from being mistaken for
unseen-plan predictability.

\subsection{Code-Agent Backend Robustness}
\label{sec:appendix_code_agent_robustness}

This experiment isolates the effect of the implementation backend. We fix the same 100 schema plans and ask seven code agents to realize them. Pass@1 records first-attempt success; failed cases receive at most one retry/repair when available. A run is counted as successful only when the generated factor passes evaluation and receives a valid reward. Signal quality is measured by mean \(|\mathrm{RankIC}|\) over final successful factors.

\begin{table}[t]
\centering
\caption{Repair-aware code-agent backend robustness on the same 100 schema plans.}
\label{tab:appendix_code_agent_robustness}
\setlength\tabcolsep{3.4pt}
\small
\resizebox{\columnwidth}{!}{%
\begin{tabular}{lccccc}
\toprule
\textbf{Backend} & \textbf{Pass@1} & \textbf{Repair} & \textbf{Final} & \(\mathbf{|RankIC|}\uparrow\) & \textbf{RankIC} \\
\midrule
GPT-5.4 & 97/100 & 2 & 99/100 & 0.0116 & 0.0040 \\
GPT-5.4-mini & 81/100 & 10 & 91/100 & 0.0129 & 0.0073 \\
Kimi-K2.7-Code & 82/100 & 10 & 92/100 & 0.0138 & 0.0075 \\
Claude Opus 4.6 & 94/100 & 3 & 97/100 & 0.0157 & 0.0113 \\
DeepSeek-V4-Flash & 63/100 & 30 & 93/100 & 0.0168 & 0.0121 \\
Qwen3.6-Flash & 49/100 & 40 & 89/100 & 0.0161 & 0.0115 \\
GLM-5.2 & 75/100 & 22 & 97/100 & 0.0145 & 0.0109 \\
\bottomrule
\end{tabular}
}
\end{table}

The corresponding visualization is reported in Figure~\ref{fig:code_agent_model_invariance} in the main paper. The results separate execution reliability from factor quality. Retry mainly improves completion rates for lower Pass@1 backends, while successful factors from different backends remain in a comparable signal range. This supports the view that the schema provides transferable implementation intent rather than relying on a single agent backend.

\section{Additional Schema Vocabulary Examples}
\label{sec:appendix_schema_examples}

We provide example schema records from each category. These examples
follow the same JSON format as the schema-to-code example in
Appendix~\ref{sec:appendix_alpha_code}.

\noindent\textbf{Event schemas.}
\begin{lstlisting}[style=appendixlisting]
{
  "schema_id": "event.range_shock_directional_move",
  "category": "event",
  "scope": "time_series",
  "name": "Range-shock directional move",
  "description": "True range or candle-body range expands materially relative to its recent level while price shows a clear directional move.",
  "terms": {
    "range shock": "an unusually large high-low, true-range, or candle-body movement.",
    "directional move": "price moves clearly upward or downward during the same event window."
  }
}

{
  "schema_id": "event.gap_fill_or_hold",
  "category": "event",
  "scope": "time_series",
  "name": "Gap fill or hold",
  "description": "After a significant opening gap, subsequent price either fills the pre-gap reference region or holds near the gap edge.",
  "terms": {
    "gap": "an opening price discontinuity relative to a previous reference price.",
    "gap edge": "the boundary of the discontinuity, such as the previous close, high, or low."
  }
}

{
  "schema_id": "event.cross_section_volume_expansion_directional_move_rank",
  "category": "event",
  "scope": "cross_sectional",
  "name": "Cross-sectional volume-expansion directional move rank",
  "description": "Ranks stocks by the strength of volume expansion accompanied by a directional price move at the same timestamp.",
  "terms": {
    "volume expansion": "trading volume is materially higher than its own recent baseline.",
    "cross-sectional rank": "relative ordering across stocks at the same timestamp."
  }
}
\end{lstlisting}

\noindent\textbf{Context schemas.}
\begin{lstlisting}[style=appendixlisting]
{
  "schema_id": "context.pivot_high_zone",
  "category": "context",
  "scope": "time_series",
  "name": "Pivot-high zone",
  "description": "Interprets the event according to whether price occurs near a recently confirmed local swing high.",
  "terms": {
    "pivot high": "a local high confirmed by a price swing.",
    "near": "a region close to the reference level, with width chosen during implementation."
  }
}

{
  "schema_id": "context.vwap_near_zone",
  "category": "context",
  "scope": "time_series",
  "name": "VWAP-near price zone",
  "description": "Interprets the event according to whether price is near its volume-weighted average price.",
  "terms": {
    "VWAP": "the volume-weighted average price.",
    "VWAP-near": "a price region close to VWAP, with width chosen during implementation."
  }
}

{
  "schema_id": "context.cross_section_premium_discount_rank",
  "category": "context",
  "scope": "cross_sectional",
  "name": "Cross-sectional premium-discount rank",
  "description": "Compares stocks by whether their price lies in a relatively premium or discount region within each stock's own reference range.",
  "terms": {
    "premium-discount": "a high or low position relative to a stock-specific reference range.",
    "cross-sectional comparison": "same-date relative ordering across stocks."
  }
}
\end{lstlisting}

\noindent\textbf{Quality schemas.}
\begin{lstlisting}[style=appendixlisting]
{
  "schema_id": "quality.volume_confirmation",
  "category": "quality",
  "scope": "time_series",
  "quality_type": "confirmation",
  "name": "Volume confirmation",
  "description": "Uses participation changes to judge whether the event is supported by external trading activity.",
  "terms": {
    "participation": "market activity measured by volume, amount, or related variables.",
    "confirmation": "an auxiliary variable supports the behavior expressed by the event."
  }
}

{
  "schema_id": "quality.path_cleanliness",
  "category": "quality",
  "scope": "time_series",
  "quality_type": "cleanliness",
  "name": "Price-path cleanliness",
  "description": "Weights the event by whether the recent price path is direct rather than repeatedly reversing.",
  "terms": {
    "price path": "the trajectory of prices within the observation window.",
    "cleanliness": "the extent to which the path has limited back-and-forth noise."
  }
}

{
  "schema_id": "quality.cross_section_volatility_confirmation_rank",
  "category": "quality",
  "scope": "cross_sectional",
  "quality_type": "confirmation",
  "name": "Cross-sectional volatility-confirmation rank",
  "description": "Compares stocks by how strongly the event is accompanied by volatility expansion at the same timestamp.",
  "terms": {
    "volatility confirmation": "the event is accompanied by range or true-range expansion.",
    "relative confirmation": "confirmation strength measured against other stocks at the same timestamp."
  }
}
\end{lstlisting}

\noindent\textbf{Direction schemas.}
\begin{lstlisting}[style=appendixlisting]
{
  "schema_id": "direction.continuation",
  "category": "direction",
  "scope": "time_series",
  "name": "Continuation direction guidance",
  "description": "Treats the event as an alpha-construction tendency in which price is expected to continue along the event-implied direction.",
  "terms": {
    "continuation": "after the event occurs, price continues along the event-implied direction.",
    "event-implied direction": "a direction obtained from breakouts, returns, deviations, or relative strength."
  }
}

{
  "schema_id": "direction.reversal",
  "category": "direction",
  "scope": "time_series",
  "name": "Reversal direction guidance",
  "description": "Treats the event as an alpha-construction tendency in which price is expected to move against the event-implied direction.",
  "terms": {
    "reversal": "after the event occurs, price moves opposite to the event-implied direction.",
    "repair or pullback": "price returns toward a reference, equilibrium, or opposite region."
  }
}

{
  "schema_id": "direction.range_oscillation",
  "category": "direction",
  "scope": "time_series",
  "name": "Range-oscillation direction guidance",
  "description": "Treats the event as insufficient for trend continuation and constructs direction from boundaries or equilibrium locations in the context.",
  "terms": {
    "oscillation": "price is more likely to rotate within a reference space than leave it unidirectionally.",
    "reference space": "a range, boundary, zone, equilibrium, or higher-timeframe location supplied by the context."
  }
}
\end{lstlisting}

\noindent\textbf{Output schemas.}
\begin{lstlisting}[style=appendixlisting]
{
  "schema_id": "output.bounded_continuous_signal",
  "category": "output",
  "scope": "time_series",
  "name": "Bounded continuous signal output",
  "description": "Compresses the directional and quality-weighted semantic score into a stable bounded final signal.",
  "terms": {
    "semantic score": "a continuous signed score formed by the event-context-quality-direction composition.",
    "bounded": "the final signal is restricted to a stable finite range."
  }
}

{
  "schema_id": "output.event_decay_signal",
  "category": "output",
  "scope": "time_series",
  "name": "Event-decay signal output",
  "description": "Keeps finite memory after a valid event trigger and gradually decays the final signal as the event becomes older.",
  "terms": {
    "finite memory": "the event affects the signal only within a limited observation window.",
    "event age": "the time elapsed since the most recent event trigger or confirmation."
  }
}

{
  "schema_id": "output.cross_section_zscore_signal",
  "category": "output",
  "scope": "cross_sectional",
  "name": "Cross-sectional z-score signal output",
  "description": "At each timestamp, demeans and standardizes the semantic score across stocks to form a relative-strength matrix.",
  "terms": {
    "demeaning": "subtracting the cross-sectional mean at the same timestamp.",
    "standardization": "dividing by the cross-sectional standard deviation and clipping extreme values."
  }
}
\end{lstlisting}

\end{document}